\documentclass[journal,transmag]{IEEEtran}
\usepackage{amsmath}
\usepackage{amsfonts}
\usepackage{amsthm,amssymb}
\usepackage{color,bm}
\usepackage{algorithmic}
\usepackage{algorithm}
\usepackage{array}
\usepackage[caption=false,font=normalsize,labelfont=sf,textfont=sf]{subfig}
\usepackage{textcomp}
\usepackage{stfloats}
\usepackage{url}
\usepackage{verbatim}
\usepackage{graphicx}
\usepackage{multirow}
\usepackage{threeparttable}
\usepackage{cite}
\newtheorem{proposition}{\bf Proposition}

\begin{document}

\title{Contour Field based Elliptical Shape Prior for the Segment Anything Model}


\author{Xinyu Zhao, Faqiang Wang\IEEEauthorrefmark{*}, Li Cui, Yuping Duan, and \IEEEauthorblockN{Jun Liu\IEEEauthorrefmark{*}}
\IEEEauthorblockA{Laboratory of Mathematics and Complex Systems (Ministry of Education of China),}
\IEEEauthorblockA{School of Mathematical Sciences, 
Beijing Normal University, Beijing, 100875, China.}
\thanks{
\IEEEauthorrefmark{*}The co-corresponding authors.
}
}



\maketitle

\begin{abstract}
The elliptical shape prior information plays a vital role in improving the accuracy of image segmentation for specific tasks in medical and natural images. Existing deep learning-based segmentation methods, including the Segment Anything Model (SAM), often struggle to produce segmentation results with elliptical shapes efficiently. This paper proposes a new approach to integrate the prior of elliptical shapes into the deep learning-based SAM image segmentation techniques using variational methods. The proposed method establishes a parameterized elliptical contour field, which constrains the segmentation results to align with predefined elliptical contours. Utilizing the dual algorithm, the model seamlessly integrates image features with elliptical priors and spatial regularization priors, thereby greatly enhancing segmentation accuracy. By decomposing the SAM into four mathematical subproblems, we integrate the variational ellipse prior to design a new SAM network structure, ensuring that the segmentation output of the SAM consists of elliptical regions. Experimental results on some specific image datasets demonstrate an improvement over the original SAM. The codes are available in \url{https://github.com/zhaoxinyum/SAM-ESP}.
\end{abstract}


\section{Introduction}
\IEEEPARstart{I}{mage} segmentation is an essential branch of computer vision with widespread applications in fields such as medical diagnosis, autonomous driving, and remote sensing. Its primary task is to accurately separate objects from the background.

With the rapid advancements in computer vision and machine learning technologies, image segmentation methods have exhibited diverse and sophisticated trends. Currently, mainstream approaches to image segmentation generally fall into two categories: variational models and deep learning-based models. Variational models, such as the Potts model \cite{potts_1952}, active contour models \cite{Kass2004SnakesAC,Chan2001ActiveCW}, and graph cut \cite{graphcut}, typically achieve image segmentation by minimizing an energy function. This energy function usually comprises two components: a fidelity term, which measures the similarity or dissimilarity between different regions in the image, and a regularization term, which helps maintain the spatial continuity and structural coherence within the segmentation results.

To further enhance the accuracy and stability of image segmentation, many variational models integrate spatial prior information. These spatial priors encompass spatial regularization, region volumes, region shapes, and domain-specific knowledge. For instance, Das \textit{et al.} \cite{compactprior} defined a non-rotationally invariant compact prior with respect to the horizontal and vertical axes. In \cite{graph_cut_connectivity}, a connectivity constraint was proposed, mandating that the pixel points provided by users must be connected to the segmented object. Veksler \cite{starshapegraphcut} proposed a star-shaped prior term, which requires that every pixel $\bm x$ on the line connecting the center $\bm c$ and any pixel $\bm y$ in the foreground lies within the foreground. Gorelick \textit{et al.}~\cite{gorelick_convexity} penalized all 1-0-1 configurations along the lines in the segmented region to constrain the convexity of the region. In~\cite{convexityYan}, a convexity constraint was expressed using the gradient of signed distance functions and integrated into the active contour model. While the convex prior is helpful for image segmentation in some fields, it is not specific enough for the task of segmenting elliptical objects.

Segmentation of circular and elliptical shapes in images presents a ubiquitous challenge across diverse domains. For instance, within medical imaging, cell nuclei and optic cups commonly display elliptical forms. Wu \textit{et al.} \cite{ellipticalshapeparametricfittingalgorithm} proposed an elliptical parameter fitting algorithm for segmenting cell images, which, to our knowledge, was the earliest attempt to integrate elliptical priors into image segmentation. This method used the image information to construct an elliptical grayscale image to approximate the original image, which was susceptible to the influence of image noise and uneven textures. Additionally, an elliptic function was integrated into the dissimilarity metric of FCM clustering by Leung \textit{et al.}~\cite{lipfuzzycluster} for lip image segmentation. Subsequently, in \cite{Graphcutelliptical}, the difference between the segmentation result and the reference ellipse mask was incorporated into the energy function of the graph cut method. This approach requires fitting the best ellipse during the iterative process to update the shape prior term, resulting in high computational costs.

In \cite{shapebaesdTomatoes,Pupilactivecontour},  authors incorporated an elliptical shape prior into the parameter active contour model. Moreover, in \cite{LVsegmentsnakemodel}, a circular energy function was integrated into the snake model for endocardium extraction. While shape prior active contour methods can enhance segmentation accuracy, they often suffer from numerical instability and are susceptible to getting trapped in local minima. 

Compared to learning-based models, variational models excel in integrating spatial prior information. However, the similarity measure in variational models is typically based on manually selected low-level image features, like grayscal intensity, color, and edges. When faced with a substantial number of supervised samples, variational models may not efficiently extract complex deep features compared to learning-based models like FCN \cite{FCN} or U-Net \cite{UNet}. However, for certain specific segmentation tasks, the available datasets are limited. In response to this challenge, researchers in recent years have been dedicated to developing universal image segmentation models, such as the Segment Anything Model (SAM) \cite{SAM}. Essentially, SAM is a promptable image segmentation model capable of accurately segmenting targets when provided with points, boxes, or text cues. SAM demonstrates robust generalization capabilities. Existing studies have successfully applied the SAM to challenging tasks \cite{med_segment,adaptsamtomed,cheng_sam-med2d,SAMoph,rsam-seg_2024}. To our knowledge, most studies apply SAM to downstream tasks by adding adapters to the network structure\cite{SAMoph,rsam-seg_2024,chen_sam_2023}, which lacks a clear mathematical basis. Due to repetitive attention layers and upsampling/downsampling in the network architecture, spatial positional information in images may be compromised, making it challenging to satisfy basic segmentation requirements like spatial smoothness and shape constraints. This defect is also common in general deep neural networks.

There are primarily three strategies for incorporating specific spatial information into the network. First, the post-processing method integrates the output features of the network into the variational model. Although this approach can improve segmentation performance to some extent, the post-processing results cannot be backpropagated during model training. Second, modifying the loss function of neural networks by introducing regularization terms can enhance network performance. Mirikharaji and Hamarneh \cite{starshapeloss} incorporated a star-shaped prior term into the cross-entropy loss. A convex shape prior term was integrated  into the loss function by Han \textit{et al.} \cite{convexshapeloss}.  Akinlar \textit{et al.} \cite{CNNpuilsegmentation} incorporated the average distance between the ground truth pupil ellipse and the detected pupil boundary as a shape prior regularization term within the loss function. However, these methods are highly sensitive to input data in the prediction since the prediction phase does not involve the loss function. Lastly, the third strategy involves unfolding the variational segmentation algorithm into corresponding modules integrated into the network structure, combining the advantages of both post-processing and loss function methods. For example, Liu \textit{et al.} \cite{DCNNwithSpatialRegularization} introduced a soft thresholding dynamic (STD) into the softmax activation function, which endows the outputs of DCNNs with specific priors such as spatial regularization, volume preservation, and star-shaped priors.

In this study, we propose a novel elliptical shape prior formulation with contour field constraints. Our work builds on the contour flow framework proposed by Chen \textit{et al.}~\cite{chen2025contourflowconstraintpreserving}, which introduces a mathematically derived constraint to preserve global shape similarity in image segmentation. By formulating this constraint from a contour flow perspective and integrating it into deep networks as a loss or via an unrolled variational model, their method improves segmentation accuracy and shape consistency across diverse datasets. We extend this foundation to explore elliptical shape-constrained segmentation. Existing methods \cite{Graphcutelliptical,shapebaesdTomatoes,Pupilactivecontour} commonly minimize the distance between the current segmentation result and a reference ellipse. In contrast, we guide the segmentation result to maintain the elliptical shape by constructing a contour field based on parameterized ellipses. Additionally, through the use of dual algorithms, we effectively integrate the elliptical prior with spatial regularization into the image segmentation model. Furthermore, we expand our variational model into a new network module, termed the ESP (Elliptical Shape Prior) module. We will demonstrate how this module can be seamlessly integrated into the Segment Anything Model (SAM)\cite{SAM}. To the best of our knowledge, there is currently no literature on the SAM architecture ensuring that the shape of the image segmentation output is elliptical.

The main contributions of this paper are as follows:
\begin{itemize}
	\item We propose an elliptical shape prior representation with the variational method by constraining the contour field.
	\item The proposed ellipse prior expression can be efficiently solved using a dual algorithm, allowing the incorporation of the ellipse prior into the design of deep network structures through a latent dual space.
	\item By decomposing SAM into mathematical sub-problems, we offer a new approach to integrate the elliptical shape prior into SAM.
\end{itemize}

The structure of this paper is outlined as follows: In Section \ref{related works}, we introduce some related works. Section \ref{section2} presents the proposed model, which incorporates constraints for elliptical shapes. Following this, in Section \ref{section3}, we introduce the ESP module and integrate it into SAM. We then demonstrate the performance of the proposed model through simple numerical experiments in Section \ref{performance of ESP}. Section \ref{numerical results} presents the experimental results of the model on different datasets. Finally, in the concluding section, we provide a summary and discussion.
\section{The Related Works}
\label{related works}

\subsection{Potts Model}
The Potts model \cite{potts_1952} is a classical variational segmentation model that encompasses other variational segmentation models within its framework. {The Potts model, after convexification and regularization \cite{Yuan2010}, can be expressed as:}
\begin{equation}\label{potts}
	\bm{u}^*=\underset{{\bm{u}\in\mathbb{U}}}{\arg\min}\left\{\begin{aligned}
		&\sum_{i=1}^I\int_{\Omega}-o_i(\bm x)u_i(\bm x)d\bm{x}\\
		&\quad+\lambda\sum_{i=1}^{I}\int_{\Omega}|\nabla u_i(\bm x)|d\bm x
	\end{aligned}
	\right\},
\end{equation}
where $\Omega$ represents the image region, and $I$ is the number of classes. $o_i(\bm x)$ denotes the similarity of pixel $\bm x$ to the $i$-th class. The second term in \eqref{potts} is the Total Variation (TV) regularization, which quantifies the boundary length of the segmentation regions. The parameter $\lambda$ modulates the trade-off between these terms, while the set
\begin{equation}\label{mathbbU}
	\mathbb{U}=\left\{\bm{u}=(u_1,\dots,u_I)\in[0,1]^I:\sum_{i=1}^Iu_i(\bm x)=1,\forall \bm x\in\Omega \right\}.
\end{equation}
stands for the image segmentation condition.

Although the TV regularization term is widely used in image segmentation, its non-smoothness, sensitivity to the choice of the parameter $\lambda$, and boundary effects limit its effectiveness.

\subsection{Threshold Dynamics Method for Regularization}
A smooth and concave regularization term is proposed in \cite{2015ThresholdDF, iterativethresholdingmethod, DCNNwithSpatialRegularization} as a replacement for the Total Variation (TV) term. This regularization term is defined as follows:
\begin{equation}\label{thresholddynamics}
\sqrt{\frac{\pi}{\sigma}}\sum_{i=1}^I{\int_\Omega u_i(\bm x)\left(k\ast (1-u_{i})\right)(\bm x)}d\bm x,
\end{equation}
where ``$\ast$" denotes convolution and $k$ is a kernel, like a 2-D Gaussian function $k(\bm x)=\frac{1}{2\pi\sigma^2}e^{-\frac{\Vert\bm x\Vert^2}{2\sigma^2}}$.

This regularization term accomplishes the task of segmenting image results into homogeneous regions by introducing penalties for the inconsistency between neighboring points and the segmentation class assigned to the central point. 
{
With this regularization, Liu \textit{et al.} \cite{DCNNwithSpatialRegularization} proposed the following Soft Threshold Dynamics (STD) regularized model:
\begin{equation}\label{Potts-std}
	\underset{{\bm{u}\in\mathbb{U}}}{\min}\left\{\langle -\bm{o},\bm{u}\rangle+\varepsilon\langle\bm{u},\ln{\bm{u}}\rangle+\lambda\langle\bm{u},k\ast(1-\bm{u})\rangle\right\}.
\end{equation}
}
Here,
 \begin{equation}\label{TD-term}
 	 \lambda\langle\bm{u},k\ast(1-\bm{u})\rangle=\lambda\sum_{i=1}^I{\int_\Omega u_i(\bm x)(k\ast (1-u_i))(\bm x)}d\bm x.
 \end{equation}
is referred to as $\mathcal{R}(\bm{u})$. Since this term is concave when $k$ is symmetric and positive semi-definite, the objective function can be optimized using the difference of convex (DC) algorithm:
\begin{equation}\label{linearation}
	\bm{u}^{t+1}=\arg\min_{\bm{u}\in\mathbb{U}}\bigl\{\langle -\bm{o}+\varepsilon\ln\bm {u},\bm{u}\rangle+\mathcal{R}(\bm{u}^t)+\langle \bm p^t,\bm{u}-\bm{u}^t \rangle\bigr\},
\end{equation}
where $\bm p^t=\lambda k\ast(1-2\bm{u}^t)\in\partial\mathcal{R}(\bm{u}^t)$. This method exhibits high stability and rapid convergence in practical implementations, in contrast to TV regularization {\cite{DCNNwithSpatialRegularization}}. Therefore, we will adopt this regularization term instead of the TV term in our model.
\subsection{Active Contour with Elliptical Shape Prior}
Active contour models, introduced by Kass \textit{et al.} \cite{Kass2004SnakesAC}, combine image data and geometric properties into an energy function. Minimizing this function allows the contour to converge to the target edge from its initial position. The general energy function consists of internal and external energy components:
\begin{equation}
	\label{active contour}
	E(\bm{v})=\int_{0}^1 E_{int}(\bm{v}(s)) + E_{ext}(\bm{v}(s))ds.
\end{equation}
Here, $\bm{v}(s)=(x(s),y(s)),s \in[0,1]$ represents the contour curve. The internal energy $E_{int}$ ensures curve continuity and smoothness, while the external energy $E_{ext}$ is generally defined by the image gradient.

To attain particular contour shapes, an additional shape energy term $E_{shape}$ is frequently integrated into \eqref{active contour}. When imposing circular constraints, this term quantifies the deviation from a reference circle \cite{LVsegmentsnakemodel}. 
\begin{equation}
	E_{shape}=\int_{0}^{1}(|\bm{v}(s)-\bm{v}_0|-R)^2ds,
\end{equation}
 where $\bm{v}_0$ and $R$ indicate the center and radius of the reference circle, respectively.
 
 For elliptical shape constraints, Verma \textit{et al.} \cite{shapebaesdTomatoes} used a polar coordinate representation, $\bm{v}(\theta)=\rho(\theta)e^{j\theta},\theta\in [0,2\pi]$ represents the contour curve, and the reference ellipse $\bm{v}_{e}=\rho_e(\theta)e^{j\theta}$. They defined the $E_{shape}$ as follows:
  \begin{equation}
  	E_{shape}=\int_{0}^{2\pi}(\rho(\theta)-\rho_{e}(\theta))^2d\theta.
  \end{equation}

To achieve a more precise fit, Ukpai \textit{et al.} \cite{Pupilactivecontour} introduced scaling, rotation, and translation transformations into the energy term.

The active contour method is widely acknowledged as an effective approach. The integration of elliptical shape priors within the active contour method has shown promise in enhancing segmentation accuracy for specific tasks\cite{shapebaesdTomatoes,Pupilactivecontour,LVsegmentsnakemodel}. However, this method's success heavily hinges upon the initial contour selection and is susceptible to convergence to local minima.
\subsection{Segment Anything Model}
SAM \cite{SAM} is a novel general-purpose visual segmentation model, which consists of three key components: an image encoder, a prompt encoder, and a mask decoder. Specifically, the image encoder, based on the Vision Transformer (ViT) \cite{vit}, maps images to a high-dimensional embedding space, denoted as $\mathcal{T}^1_{\theta_1}$. The prompt encoder converts positional prompts to 256-dimensional embeddings, denoted as $\mathcal{T}^2_{\theta_2}$. The mask decoder employs two transformer modules to merge image embeddings with prompt features and uses transposed convolution layers for upsampling, followed by MLP layers to map the output tokens to a dynamic classifier.

To combine with the elliptical shape prior proposed in this paper later, we first express SAM as mathematical sub-problems. Let ${I{mg}}\in\mathbb{R}^{h\times w}$ denotes the input image and $P\in\mathbb{R}^{k}$ represents the input prompt with $k$ tokens. 
The parameterized encoding operators $\mathcal{T}^1_{\theta_1}: ~\mathbb{R}^{h\times w}\rightarrow \mathbb{R}^{h_1\times w_1\times c_1}$ and $\mathcal{T}^2_{\theta_2}:~\mathbb{R}^{k}\rightarrow \mathbb{R}^{k\times c_1}$ respectively map the image and prompt to features. To integrate features from these two different spaces, a fusion operator $\mathcal{T}^3_{\theta_3}:~\mathbb{R}^{h_1\times w_1\times c_1}\times \mathbb{R}^{k\times c_1}\rightarrow \mathbb{R}^{h\times w}$ would be introduced. Operator $\mathcal{T}^3_{\theta_3}$ includes cross-attention, MLP and upscale opteration. Then we can write the SAM as:
\begin{equation}\label{samdecoder}
	\begin{cases}
		{\widetilde{Img}} = \mathcal{T}^1_{\theta_1}({Img}) ,
		\widetilde{P} = \mathcal{T}^2_{\theta_2}(P), \\
		o = \mathcal{T}^3_{\theta_3}({\widetilde{Img}},\widetilde{P}), \\
		u = \mathcal{F}(o).
	\end{cases}
\end{equation}
Here $o$ is the extracted feature by SAM, and $\mathcal{F}$ is a decoding operator. In the simplest case, $\mathcal{F}$ can be the Heaviside step function defined as follows:
{
\begin{equation}\label{Heaviside}
	\mathcal{F}(o)( {\bm{x}})=\left\{
	\begin{array}{rcl}	
		0 ,& &{o( {\bm{x}})\leq 0},\\
		1 ,& &{o( {\bm{x}}) > 0}.
	\end{array}
	\right.		
\end{equation}}
 {The backward propagation derivative of this non-differentiable function at $0$ is usually handled with special treatment by automatic differentiation in PyTorch and TensorFlow.}

The final output of SAM is given by $u\in [0,1]^{h\times w}$. 
In summary, we represent SAM as the above four sub-problems: where $\mathcal{T}^1_{\theta_1}$, $\mathcal{T}^2_{\theta_2}$ are encoding processes, $\mathcal{T}^3_{\theta_3}$ primarily involves the cross attention feature fusion step, and $\mathcal{F}$ is the decoding process.

To apply SAM to downstream tasks, Chen \textit{et al.} \cite{chen_sam_2023} introduced SAM-adapter, which consists of only two MLP layers inserted between the transformer layers of the image encoder. This integration enables the model to learn task-specific information.

For adapting SAM to retinal image segmentation, Qiu \textit{et al.} proposed a modification \cite{SAMoph} where they fixed the parameters of the image encoder and introduced a learnable prompt layer between each transformer layer. This prompt layer incorporates convolutional layers, layer normalization, and GELU activation. Additionally, they replaced the original prompt encoder and mask decoder with a trainable task head composed of convolutional and linear layers.

Similarly, Zhang \textit{et al.} \cite{rsam-seg_2024} devised an Adapter-Scale comprising Downscale, ReLU, and Upscale layers after the partial multi-head self-attention blocks in SAM's image encoder. Additionally, they added an Adapter-Feature, constituted by two MLP layers, between transformer layers to merge high-frequency image information. These methods all involve augmenting SAM's image encoder with learnable network layers, termed adapters. However, the rationale behind selecting these adapters lacks explicit mathematical justification. 

 {While existing studies\cite{cohen2025texturesam,WANG2024110685,zhang2025,SAM_Texture} have examined SAM's segmentation characteristics, its core perceptual preferences (such as texture versus shape) remain under investigation with no clear consensus yet. This motivates our work to introduce explicit elliptical shape priors to guide SAM for targeted segmentation tasks.}
In this paper, we primarily focus on the decoding process, aiming to integrate the ellipse prior into this process. It is worth noting that when $\mathcal{F}$ is the Heaviside step function,  {it is equivalent to} the minimization problem 
\begin{equation}\label{subpro4}
\mathcal{F}(o)=\underset{u\in[0,1]}{\arg\min}\{\langle-o,u\rangle\}.
\end{equation}
 {The optimization problem can be approximated by the following entropic regularization
\begin{equation}
	\label{smoothencoder}
	u_{\varepsilon}=\underset{u\in[0,1]}{\arg\min}\left\{\langle-o,u\rangle+\varepsilon\left(\langle u,\ln{u}\rangle+\langle 1-u,\ln{(1-u)}\rangle\right)\right\}.
\end{equation}
This smoothed formulation provides a continuous optimization basis for integrating the elliptical shape prior constraint in our subsequent model.}
 
\section{The Proposed Method}
\label{section2}
The objective of image segmentation is to obtain a segmentation function $u$ that is consistent with the ground truth. In this section, we use elliptical contour field to mathematically impose elliptical shape constraint on the segmentation function $u$ and incorporate this constraint into the image segmentation model.
\subsection{Elliptical Shape Prior based on Contour Flow}
{
The geometric characteristics of an ellipse are defined by its center point {$(x_0,y_0)$}, major and minor axes {$a$, $b$}, and a rotation angle {$\theta$} about the center. An ellipse with parameter $\Lambda:=(x_0, y_0, a, b, \theta)$ can be represented as follows:
\begin{align*}
	&\frac{\left((x-x_0)\cos{\theta}+(y-y_0)\sin{\theta}\right)^2}{a^2} \\
	&\quad + \frac{\left(-(x-x_0)\sin{\theta}+(y-y_0)\cos{\theta}\right)^2}{b^2} = 1.
\end{align*}
}

For the convenience of subsequent expressions,  let us represent the above ellipse equation in parametric form
\begin{equation}\label{ell-equ}
	\begin{cases}
			x= \phi(t) := a\cos\theta\cos t - b\sin\theta\sin t + x_0, \\
			y= \psi(t) := a\sin\theta\cos t + b\cos\theta\sin t + y_0,
	\end{cases}
\end{equation}
where $ t\in(0,2\pi]$. Then the tangent vector of the ellipse can be written as  {$\bm{T}_{\Lambda}(x,y)=~(\phi^{'}(t),\psi^{'}(t))$}. 

{
In the Cartesian coordinate system, $\bm{T}_{\Lambda}$ is formulated as:
\begin{equation}\label{T:eq1}
	\begin{aligned}
		\bm{T}_{\Lambda}(x,y) = \big(&\begin{aligned}[t]
			&(x-x_0)(b^2-a^2)\cos{\theta}\sin{\theta} \\
			&+(y-y_0)(b^2\sin^2{\theta}+a^2\cos^2{\theta}),
		\end{aligned} \\
		&\quad (y-y_0)(a^2-b^2)\cos{\theta}\sin{\theta} \\
		&-(x-x_0)(a^2\sin^2{\theta}+b^2\cos^2{\theta})\big).
	\end{aligned}
\end{equation}
}
{
To ensure that the segmentation results exhibit elliptical shapes, we do not impose direct constraints on $u(x,y)$. Inspired by Chen \textit{et al.} \cite{chen2025contourflowconstraintpreserving}, we enforce an orthogonality constraint between its gradient $\nabla u(x,y)$ and a tangent vector field $\bm{T}_{\Lambda}(x,y)$ determined by a given parameter $\Lambda$. In fact, it can be proven that if $\langle\nabla u(x,y), \bm{T}_{\Lambda}(x,y)\rangle=0, \forall (x,y)\in\Omega$, then the contours of $u$ must be ellipses, as stated in the following proposition.} 
{
\begin{proposition}\label{proposition1}
	Suppose both $u: \Omega\subset\mathbb{R}^2\rightarrow[0,1]$ and its contours be $\mathcal{C}^1$, and $\bm{T}_{\Lambda}(x,y)$ is the tangent vector field  determined by \eqref{T:eq1}. Then we have
	all the contours of $u(x,y)$ are concentric ellipses with the same orientation if and only if $\exists\Lambda, s.t.\langle\nabla u(x,y), \bm{T}_{\Lambda}(x,y)\rangle=0, \forall (x,y)\in\Omega$.
\end{proposition}
Please find the proof in the Appendix \ref{appendix_A}.
}
Now, let us define
$$\mathbb{P}_{\Lambda}=\{u: \langle \nabla u(x,y),\bm{T}_{\Lambda}(x,y)\rangle =0, \forall (x,y)\in\Omega \}.$$ 
According to the Proposition \ref{proposition1}, if $u\in\mathbb{P}_{\Lambda}$, then all the contours of $u$ must be ellipses.
Please note that $\mathbb{P}_{\Lambda}$ is a convex set  {for a given $\Lambda$}, making it easy to satisfy the constraint $u\in\mathbb{P}_{\Lambda}$ using the dual method in the variational problem. Therefore, we can propose an image segmentation model with our elliptical shape prior as follows.

\subsection{Our Model with Elliptical Shape Prior}
 {If the $i$-th segmentation target needs to maintain an elliptical shape, then our model can be given as}:
 {
\begin{equation}\label{ellipsemodel}
	\begin{aligned}	
    &\underset{\bm{u}\in\mathbb{U},{\Lambda}}{\min}~\left\{\langle-\bm{o},\bm{u}\rangle+\varepsilon\langle\bm{u},\ln(\bm{u})\rangle+\mathcal{R}(\bm{u})\right\},\\
    &s.t.\ \nabla u_i(x,y)\cdot \bm{T}_{\Lambda}(x,y)=0,\forall (x,y)\in\Omega 
		\end{aligned}
\end{equation}}
where $u_i$ is the segmentation function of $i$-th region, $\mathcal{R}(\bm{u})$ denotes the threshold dynamic regularization term \cite{DCNNwithSpatialRegularization} mentioned earlier in \eqref{TD-term}. 

The above model differs from the existing model  \eqref{Potts-std} in three main aspects: Firstly, by imposing $u_i\in\mathbb{P}_{\Lambda}$, we constrain the $i$-th segmentation region to satisfy the elliptical shape prior. Secondly, in order to satisfy the smoothness condition of $u_i$ mentioned in Proposition \ref{proposition1}, we add the second term as an entropy regularization term, in which the parameter $\varepsilon>0$ controls the smoothness of $u_i$. If $\varepsilon\rightarrow 0^+$, then the segmentation function $\bm{u}$ tends to be binary.
Thirdly, we need to estimate the tangent vector field of the prior ellipses, which enables the model to find the optimal elliptical segmentation region.

 {According to the Lagrange multiplier method, we transform the constrained problem into the following saddle-point formulation}:
\begin{equation}\label{dual}
	\min_{\bm{u}\in\mathbb{U},  {\Lambda}}\max_{q}\bigl\{\langle-\bm{o},\bm{u}\rangle+\mathcal{R}(\bm{u})+\varepsilon\langle\bm{u},\ln(\bm{u})\rangle-\langle q,\bm{T}_{ {\Lambda}}\cdot\nabla u_{i}\rangle\bigr\}, 
\end{equation}
where the variable $q(x,y)$ represents the dual variable (Lagrange multiplier). By using the conjugate formula, we obtain: $\langle q,\bm{T}_\Lambda\cdot\nabla u_{i}\rangle=-\langle \text{div}(q\bm{T}_\Lambda), u_{i}\rangle$  {with the Dirichlet boundary condition $u_i|_{\partial\Omega}=0$}. Here ``div" is the divergence operator which is the conjugate operator of the gradient. Thus, we can rewrite \eqref{dual} as:
\begin{equation}\label{dual2}
	\min_{\bm{u}\in\mathbb{U},  {\Lambda}}\max_{q}\left\{
		\underbrace{\begin{aligned}
				&\langle-\bm{o},\bm{u}\rangle+\mathcal{R}(\bm{u})+\varepsilon\langle\bm{u},\ln(\bm{u})\rangle \\
				&\quad+\langle \text{div}(q\bm{T}_{ {\Lambda}}),u_{i}\rangle
		\end{aligned}}_{\mathcal{E}(\bm u, q,  {\Lambda)}}
	\right\}.
\end{equation}

 {Inspired by dual ascent method\cite{luo1993convergence,2017Accelerated}, we utilize the following alternating update strategy and obtain three subproblems:}
\begin{equation}\label{threesub}
	\begin{cases}
		\begin{aligned}
            q^{t+1}&=\underset{q}{\arg\max}~\mathcal{E}(\bm u^{t}, q,  {\Lambda^t}) {-\frac{1}{2\tau_q}\Vert q-q^t\Vert^2},\\
			\bm{u}^{t+1} &= \underset{\bm{u}\in\mathbb{U}}{\arg\min}~\mathcal{E}(\bm u, q^{t+1},  {\Lambda^t}), \\
			 {\Lambda^{t+1}} &= \underset{\Lambda}{\arg\min}~\mathcal{E}(\bm u^{t+1}, q^{t+1},  {\Lambda}).
		\end{aligned}
	\end{cases}
\end{equation}

 {The first subproblem associated with variable $q$ is equivalent to the following equation:}

\begin{equation}\label{q:iter}
	q^{t+1}=q^t-\tau_q\bm{T}_{ {\Lambda^t}}\cdot\nabla u_{i}^t,
\end{equation}
where $\tau_q>0$ is a step parameter. 

For the second subproblem related to $\bm u$, we can employ the DC (Difference of Convex functions) algorithm. By replacing $\mathcal{R}(\bm{u})$ with its supporting hyperplane \cite{DCNNwithSpatialRegularization}, one can get:
\begin{equation*}
	\bm{u}^{t+1}=\underset{\bm{u}\in\mathbb{U}}{\arg\min}\left\{
	\begin{aligned}
	&\langle-\bm{o}+\bm{p}^t,\bm{u}\rangle+\varepsilon\langle\bm{u},\ln(\bm{u})\rangle\\
	&\quad+\langle \text{div}(q^{t+1}\bm{T}_{ {\Lambda^t}}),u_{i}\rangle
		\end{aligned}
	\right\}.		
\end{equation*}
Here $\bm{p}^t=\lambda k\ast(1-2\bm{u}^t)\in\partial\mathcal{R}(\bm{u}^t)$ is a subgradient of $\mathcal{R}(\bm u)$ as shown in \eqref{linearation}. This $\bm{u}$-subproblem 
is a strictly convex optimization problem, guaranteeing the existence of a unique solution. Furthermore, this solution takes the form of a closed-form softmax expression:
\begin{equation}\label{u:iter}
	{u}^{t+1}_{\hat{i}}=\frac{e^{\frac{\bm{o}_{\hat{i}}-p_{\hat{i}}^t-\delta_{\hat{i},i}\text{div}(q^{t+1}\bm{T}_{\Lambda^t})}{\varepsilon}}}{\sum_{i^{'}}^{I}e^{\frac{\bm{o}_{i^{'}}-p_{i^{'}}^t-\delta_{i^{'},i}\text{div}(q^{t+1}\bm{T}_{\Lambda^t})}{\varepsilon}}},~~\hat{i}=1,\dots,I.
\end{equation}
Here $\delta_{\hat{i},i}$ is the Delta function defined by
$\delta_{\hat{i},i} =0$ when $\hat{i}\neq i$ and $\delta_{i,i} =1$. {More details for calculating $u^{t+1}$ can be found in Appendix \ref{appendix_B}.}

{
Now let us solve the third subproblem. First, we note that $\bm T_\Lambda$ is determined by parameter $\Lambda=(x_0, y_0, a, b,\theta)$. 
Directly solving this subproblem to update the parameter $\Lambda$ entails a substantial computational burden. In \cite{lipfuzzycluster}, the authors utilized the second-order moment of the membership function to obtain the relevant parameters of the best-fitting ellipse. We follow the same approach. In fact, if $u_i$ denote the indicator function representing an elliptical region. Then we have the following formulas:}
\begin{equation}\label{approx}
	\begin{cases}
			\iint_{\Omega}u_{i}(x,y)dxdy=\pi ab,\\ \iint_{\Omega}xu_{i}(x,y)dxdy= x_0\pi ab, \\
			\iint_{\Omega}yu_{i}(x,y)dxdy= y_0\pi ab, \\
			\iint_{\Omega}(x-x_0)^2u_{i}(x,y)dxdy=\frac{(a^2\cos^2{\theta}+{b^2}\sin^2{\theta})}{C}, \\
			\iint_{\Omega}(y-y_0)^2u_{i}(x,y)dxdy=\frac{(a^2\sin^2{\theta}+b^2\cos^2{\theta})}{C}, \\
			\iint_{\Omega}(y-y_0)(x-x_0)u_{i}(x,y)dxdy=\frac{(a^2-b^2)\cos{\theta}\sin{\theta}}{C}.
	\end{cases}
\end{equation}
{where $C=\frac{4}{ab\pi}$.
Since we only need to update $\bm{T}_{ {\Lambda}}$, based on equation \eqref{approx}, we obtain the following scheme to update the important quantities related to $\bm{T}_{ {\Lambda^{t+1}}}$:
\begin{equation}\label{approellipse}
	\begin{cases}
		x^{t+1}_0=\frac{\sum\limits_{x,y}xu_{i}^{t+1}(x,y)}{\sum\limits_{x,y}u_{i}^{t+1}(x,y)},y^{t+1}_0=\frac{\sum\limits_{x,y}yu_{i}^{t+1}(x,y)}{\sum\limits_{x,y}u_{i}^{t+1}(x,y)},\\
		((a^{t+1})^2\!-\!(b^{t+1})^2)\cos{\theta^{t+1}}\sin{\theta^{t+1}}=\frac{4M_{xy}}{\sum\limits_{x,y}u_{i}^{t+1}(x,y)},\\
		(b^{t+1})^2\sin^2{\theta^{t+1}}\!+\!(a^{t+1})^2\cos^2{\theta^{t+1}}\!=\!\frac{4M_{xx}}{\sum\limits_{x,y}u_{i}^{t+1}(x,y)},\\
		(a^{t+1})^2\sin^2{\theta^{t+1}}\!+\!(b^{t+1})^2\cos^2{\theta^{t+1}}\!=\!\frac{4M_{yy}}{\sum\limits_{x,y}u_{i}^{t+1}(x,y)}.
	\end{cases}
\end{equation}
where 
$
M_{xx} = \sum\limits_{x,y} (x - x^{t+1}_0)^2 u_i^{t+1}(x,y),
M_{yy} = \sum\limits_{x,y} (y - y^{t+1}_0)^2 u_i^{t+1}(x,y),
M_{xy} = \sum\limits_{x,y} (x - x^{t+1}_0)(y - y^{t+1}_0) u_i^{t+1}(x,y)
$ represent the second moments.
}
Once we obtain the values for the parameters mentioned above, we can update $\bm T_{ {\Lambda^{t+1}}}$ according to equation \eqref{T:eq1}.

In summary, Algorithm \ref{algorithm} outlines the steps to solve the proposed model.

\begin{algorithm}[!t]
	\caption{Image segmentation with elliptical shape prior}
	\label{algorithm}
	\begin{algorithmic}
		\REQUIRE The image feature $\bm{o}$.
		\ENSURE Segmentation function $\bm{u}$, where $u_i$ represents the indicator function for the ellipse region. 
		\STATE \textbf{Initialization:} $u^0_{\hat i}=[softmax(\bm o)]_{\hat i}=\frac{e^{\frac{o_{\hat{i}}}{\varepsilon}}}{\sum_{i^{'}}^{I}e^{\frac{o_{i^{'}}}{\varepsilon}}}$, dual variable $q^0=0$, and a tangent vector field $\bm T_{\Lambda^0}$.
		\FOR{$t=0,1,2,\dots$}
		\STATE \textbf{ 1.} Solve the first subproblem to update variable $q^{t+1}$ according to (\ref{q:iter}):
		\[
		q^{t+1}=q^t-\tau_q\bm{T}_{\Lambda^t}\cdot\nabla u_{i}^t.
		\]
		\STATE \textbf{ 2.} Solve the second subproblem using the DC algorithm to update variable $\bm u^{t+1}$ according to (\ref{u:iter}):
		\[
		\bm{u}^{t+1}_{\hat{i}}=softmax(\bm{o}_{\hat i}-\bm{p}_{\hat i}^t-\delta_{\hat{i},i} \text{div}(q^{t+1}\bm{T}_{\Lambda^t})).
		\]
		\STATE \textbf{ 3.}  {Calculate the elliptical parameter $\Lambda^{t+1}$ via (\ref{approellipse}) then update $\bm T_{\Lambda^{t+1}}$ according to (\ref{T:eq1})}.
		\STATE \textbf{ 4.} Convergence check. If it  {is} converged, end the algorithm.
		\ENDFOR
	\end{algorithmic}
\end{algorithm}
\section{Integrated into the Segment Anything Model}
\label{section3}
In this section, we will unroll the alternating iterative process of Algorithm \ref{algorithm} into a neural network architecture, then integrate it into the SAM.
\subsection{Elliptical Shape Prior (ESP) Module}
\begin{figure*}[!t]
\centering
	\includegraphics[width=0.85\textwidth]{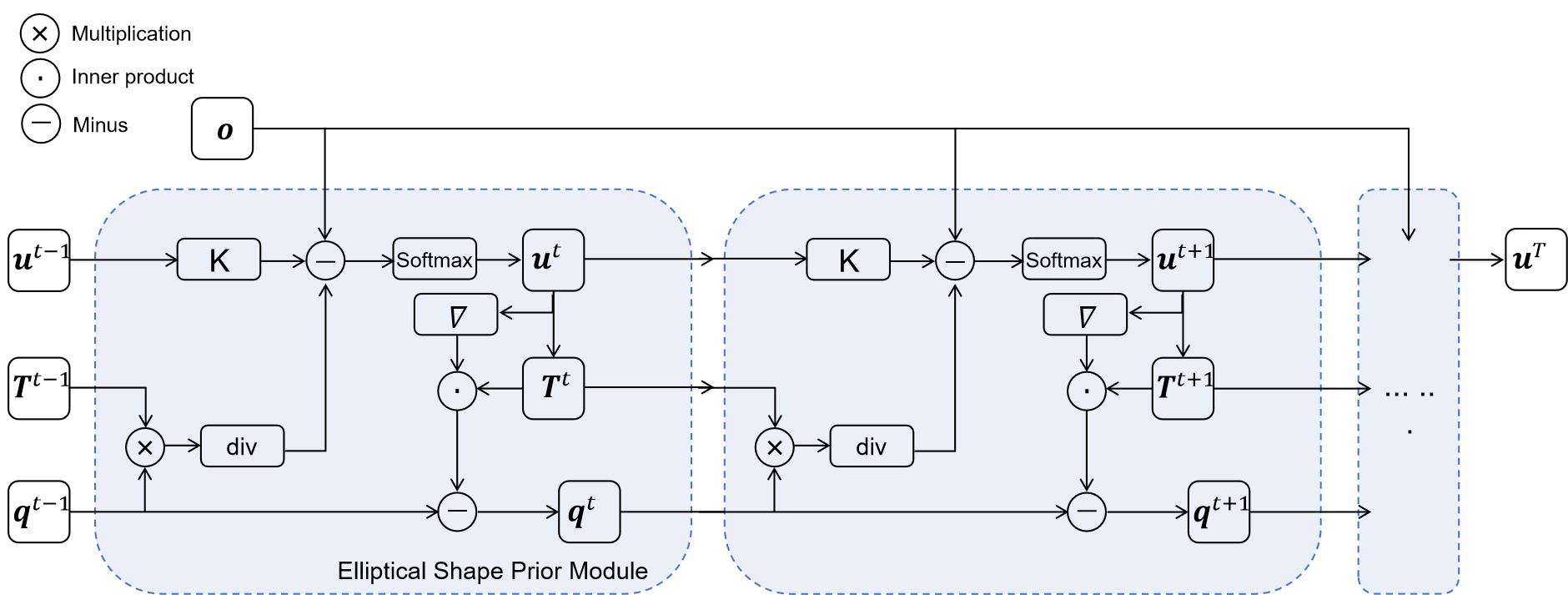}
	\caption{Architecture of each elliptical shape prior (ESP) module.}
	\label{ESP}      
\end{figure*}

We incorporate our elliptical constraint into equation (\ref{smoothencoder}) and replace the final subproblem of the SAM (\ref{samdecoder}) with this variational problem, yielding 
\begin{equation}\label{samesp}
		\begin{cases}
			 {\widetilde{Img}} = \mathcal{T}^1_{\theta_1}( {Img}) ,\quad
		\widetilde{P} = \mathcal{T}^2_{\theta_2}(P), \\
		o = \mathcal{T}^3_{\theta_3}( {\widetilde{Img}},\widetilde{P}), \\
         {
			u^*,\Lambda^*= \underset{u\in[0,1],\Lambda}{\arg\min}\left\{
			\begin{aligned}
				&\langle-o,u\rangle+\mathcal{R}(u)+\raisebox{0.5ex}{$\chi$}_{\mathbb{P}_{\Lambda}}(u)+\\
				&\varepsilon(\langle u,\ln{u}\rangle+\langle 1-u,\ln{(1-u)}\rangle)
			\end{aligned}
			\right\}.
            }
		\end{cases}
\end{equation}
The third sub-equation in \eqref{samesp} represents the scenario of our model \eqref{ellipsemodel} in binary segmentation.  {Here $\raisebox{0.5ex}{$\chi$}_{\mathbb{P}_{\Lambda}}$ represents the indicator functional of the set $\mathbb{P}_{\Lambda}$, meaning it equals $0$ when $u\in\mathbb{P}_{\Lambda}$, and $+\infty$ otherwise.}
This ensures that the decoder of SAM produces smooth elliptical object.

We unroll the Algorithm \ref{algorithm} into several network layers to bridge the gap between optimization algorithms and deep neural architectures. 
Specifically, the operators $\nabla$ and ``div'' in the alternating iterative scheme are replaced with fixed kernel convolution layers representing discrete gradient and divergence, respectively. According to Algorithm \ref{algorithm}, each ESP module outputs an intermediate segmentation function $u^t$, which is used to compute $u^{t+1}$ for the subsequent module. The specific structure of the ESP module is illustrated in \figurename\ref{ESP}.  For simplification, we call SAM
with this decoding module as SAM-ESP.

\section{The Performance of ESP Module}
\label{performance of ESP}
In this section, we design two very simple experiments to show the intuition of the proposed Algorithm \ref{algorithm}.

\subsection{Synthetic Image Experiment}
\begin{figure}[!t]
	\centering  
	\subfloat[{\small Image}]{
		\label{Fig.sub.1}
		\includegraphics[width=0.1\textwidth]{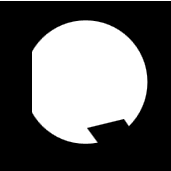}}
	\hfil
	\subfloat[{\small Iter 50}]{
		\label{Fig.sub.2}
		\includegraphics[width=0.1\textwidth]{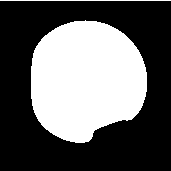}}
	\hfil
	\subfloat[{\small Iter 100}]{
		\label{Fig.sub.3}
		\includegraphics[width=0.1\textwidth]{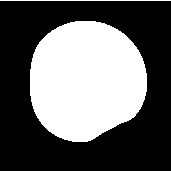}}
	\hfil
	\subfloat[{\small Iter 500}]{
		\label{Fig.sub.4}
		\includegraphics[width=0.1\textwidth]{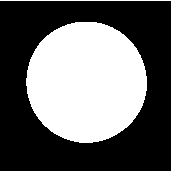}}
	\caption{The proposed method for a non-elliptical region.}
	\label{non-ellitical}
\end{figure}

Initially, our model is tested on a non-elliptical synthetic image.
For this experiment, we choose the region variance as the similarity $o_i(x)=-\|h(x)-m_{i}\|^2$ for $i=1,2$, where $h(x)$ denotes the intensity of pixel $x$. Here $m_{1}$ and $m_{2}$ represent the average gray value of the foreground and background, respectively. 
As shown in \figurename \ref{non-ellitical}, after 500 iterations, our model successfully transformed the non-elliptical region into an ellipse.

Throughout the iteration process, our model ensures orthogonality between the normal vector field of the image segmentation results and the tangent vector field of the ellipses. This alignment facilitates the transformation of the regions into elliptical shapes.

\subsection{Natural Image Segmentation}
\begin{figure}[!t]
	\centering
	\subfloat[K-means]{
		\includegraphics[width=1in]{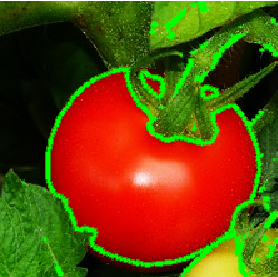}
	}\hfil
	\subfloat[{STD\cite{DCNNwithSpatialRegularization}}]{
		\includegraphics[width=1in]{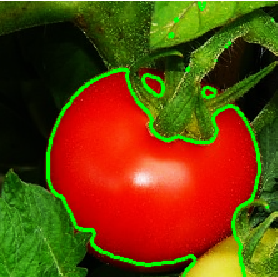}
	}\hfil
	\subfloat[{SSTD\cite{DCNNwithSpatialRegularization}}]{
		\includegraphics[width=1in]{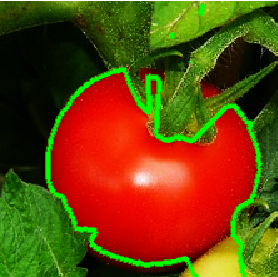}
	}\hfil
	\\
	\subfloat[{CV-CSP\cite{convexityYan}}]{
		\includegraphics[width=1in]{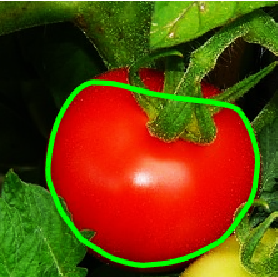}
	}\hspace{0.15cm}
	\subfloat[Proposed ESP]{
		\includegraphics[width=1in]{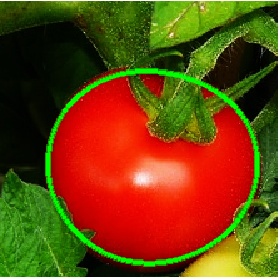}
	}%
	\caption{The comparison of different methods on a natural image.}
	\label{natureimage}
\end{figure}
The second experiment evaluates the algorithm's performance on natural images. We select an image of a tomato and compare the result with segmentation models that incorporate different shape priors: star-shaped prior (SSTD)\cite{DCNNwithSpatialRegularization} and convexity prior (CV-CSP)\cite{convexityYan}. 

In this numerical experiment, the similarity is defined as
\begin{equation*}
    o_i(x) = -(\bm{h}(x)-\bm{m}_i)^\top\bm{\Sigma}_{i}^{-1}(\bm{h}(x)-\bm{m}_i),
\end{equation*}
for $i=1,2$. Here $\bm{m}_i$ and $\bm{\Sigma}_{i}$ are the mean and covariance matrix of object and background, respectively. The convergence standard is $\|\bm{u}^{t}-\bm{u}^{t+1}\|<3\times 10^{-5}$. For SSTD\cite{DCNNwithSpatialRegularization}, we replace the center of the star-shaped regions with their centroids.

The comparison results are shown in \figurename\ref{natureimage}. Compared to K-means, both STD\cite{DCNNwithSpatialRegularization}, SSTD\cite{DCNNwithSpatialRegularization}, and CV-CSP\cite{convexityYan} methods provide smooth segmentation boundaries due to the presence of regularization terms. However, due to occlusion by green leaf stems, they fail to extract the entire tomato. In contrast, our algorithm effectively segments the tomato. This straightforward numerical illustration highlights the effectiveness of the proposed algorithm in segmenting elliptical regions.

\section{Numerical Experimental Results of SAM-ESP}
\label{numerical results}
In this section, we train SAM-ESP on several datasets and compare the experimental results with the original SAM, fine-tuned SAM, and two other methods of incorporating priors into neural network architectures, namely post-processing and loss function modification. The experimental results demonstrate that SAM-ESP achieves the best segmentation outcomes and exhibits greater stability. For experiments based on other networks, please refer to Appendix \ref{appendix_D}.

\subsection{Training Protocol and Experimental Setup}
SAM offers three different scales of image encoders: ViT-B, ViT-L, and ViT-H. ViT-L and ViT-H only have slight improvements over ViT-B but require significantly more computation \cite{SAM}. Therefore, we choose ViT-B as the image encoder.

\begin{figure*}[htbp]
\centering
	\includegraphics[width=0.8\textwidth]{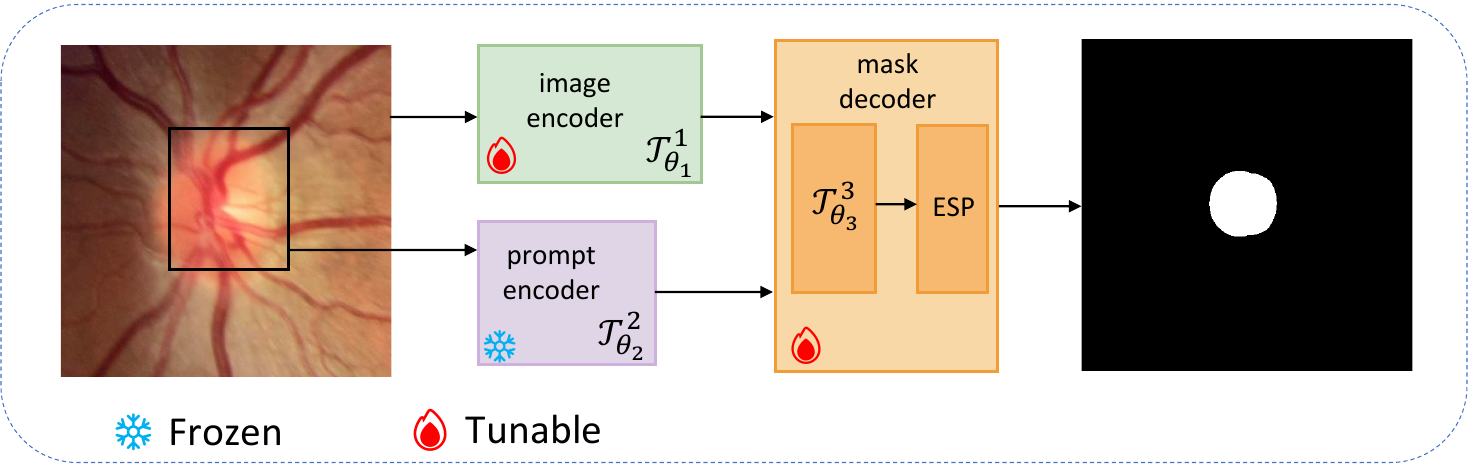}
	\caption{The network architecture of SAM-ESP}
	\label{SAM-ESPpicture}      
\end{figure*}

We maintain the core architecture of SAM. During training, all trainable parameters within the image encoder and mask decoder are updated while the prompt encoder is frozen as it proficiently encodes bounding box prompts. The detailed architecture of the entire network is illustrated in \figurename\ref{SAM-ESPpicture}. Throughout the training process, each image is provided with bounding box prompts derived from the minimum bounding box containing the ground truth, with random perturbations ranging from 0 to 40 pixels. The entropy parameter $\varepsilon$, regularization parameter $\lambda$, and gradient update rate $\tau_q$ in the ESP module are all set to 1. And we use a 5$\times$5 Gaussian kernel with a fixed standard deviation $\sigma=5$.
{Refer to Appendix \ref{appendix_C} for the discussion on hyperparameter impacts.} A total of 100 ESP modules are deployed. 

The loss function employed during training is the cross-entropy loss $\mathcal{L}_{CE}$. We utilize the Adam optimizer with default parameters. The training is implemented on an NVIDIA Tesla V100 GPU.  {Due to GPU memory constraints imposed by our hardware performance limitations, the batch size for all data is set to 1.} 
 {For each dataset, we performed three independent training runs with distinct random seeds. During each run, the model was trained over multiple epochs, and we selected the checkpoint corresponding to the highest validation Dice score for testing rather than the final epoch checkpoint to avoid overfitting.}
For comparison, we fine-tuned SAM using the same training strategy, which we refer to as SAM-fine.

To better compare the incorporation of the elliptical prior into the SAM model in different ways, we attempt another two different ways.
The first method is the post-processing method.
We use the output feature $\bm o$ of SAM-fine as the input for Algorithm \ref{algorithm}, iterating the algorithm 100 times to obtain the elliptical shape. We denote this as SAM-post. In this approach, the elliptical prior cannot guide the learning of the main network through backpropagation.

The second method of incorporating priors into the network, namely the loss function method. We can design a shape loss $\mathcal{L}_S$  based on cosine distance, as provided by Proposition \ref{proposition1}, to impose shape constraints:
\begin{equation}
	\mathcal{L}_S:=\sum_{x\in\Omega}\frac{\vert\nabla \bm{u}(x)\cdot\bm{T}_{ {\Lambda}}(x)\vert}{\|\nabla\bm{u}(x)\|}.
\end{equation}
Here, $\bm{T}_{\Lambda}$ represents the tangent vector field of the ellipse. Considering that the ground truth in the selected dataset are ellipses, we directly employ the contour flow field \cite{chen2025contourflowconstraintpreserving} of the ground truth as $\bm{T}_{\Lambda}$. By adding this shape loss term to the cross-entropy loss, the final loss function is defined as:
\begin{equation*}
	\mathcal{L}=\lambda_{l}\mathcal{L}_{CE}+(1-\lambda_{l})\mathcal{L}_{S}.
\end{equation*}
We utilize this loss to train SAM, with the parameter $\lambda_{l}$ set to 0.4 during training. The model trained with this loss function is denoted as SAM-Sloss. In this method, the elliptical prior has limited influence during the prediction process, and small input perturbations in the network can render the elliptical prior ineffective.
\subsection{Datasets}
\begin{figure}[htbp]
	\centering
	\includegraphics[width=0.1\textwidth]{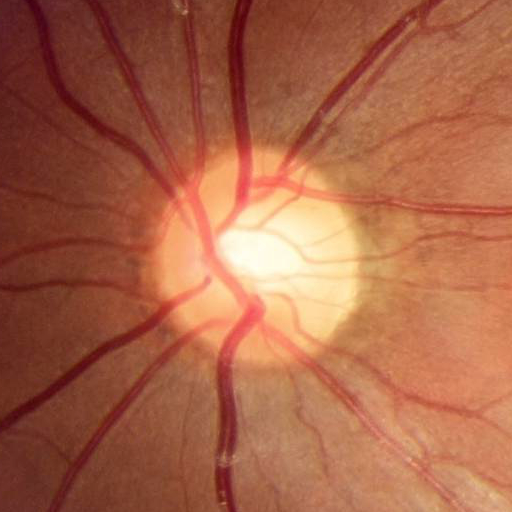}
	\includegraphics[width=0.1\textwidth]{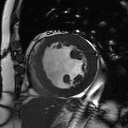}
	\includegraphics[width=0.1\textwidth]{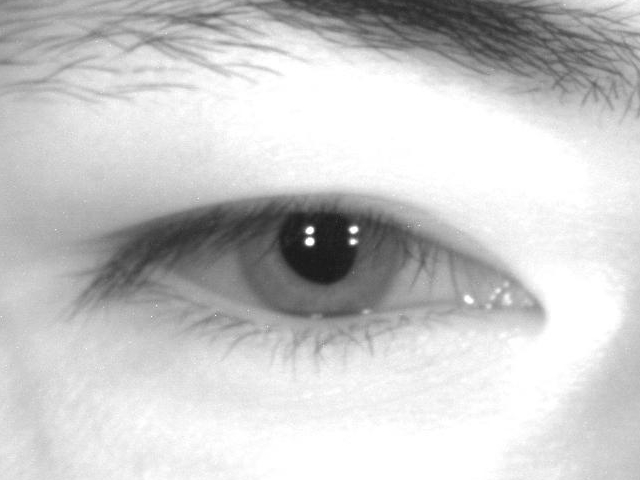}
	\includegraphics[width=0.1\textwidth]{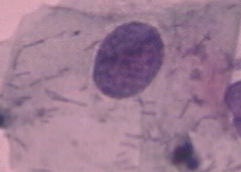}\\
	\vspace{0.1cm}
	\includegraphics[width=0.1\textwidth]{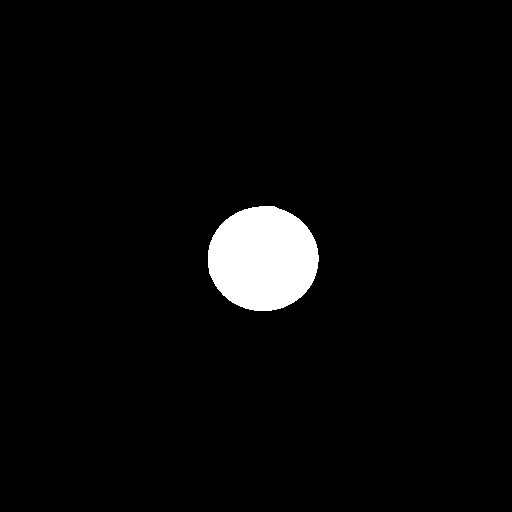}
	\includegraphics[width=0.1\textwidth]{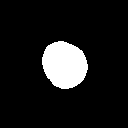}
	\includegraphics[width=0.1\textwidth]{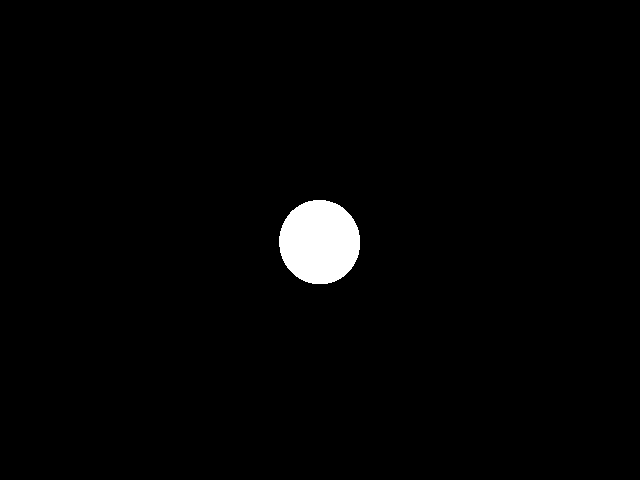}
	\includegraphics[width=0.1\textwidth]{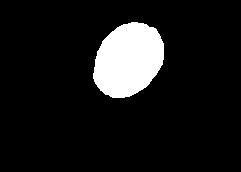}

	\caption{Sample images from the datasets used. From left to right, columns one to four are from REFUGE, ACDC, CASIA.v4-distance, DTU/Herlev dataset respectively. The first row shows the images, whose ground truth segmentations displayed at the second row}
	\label{samples}
\end{figure}

The proposed method is primarily validated and evaluated on datasets consisting of elliptical-shaped objects. The datasets selected ranges from 98 to 1200 images, each with varying resolutions. \figurename\ref{samples} illustrates sample images from the datasets, while Table \ref{dataset data} presents the data preprocessing details for the datasets used. 
\begin{table}[!t]
	\caption{Data preprocessing of the datasets used}
	\label{dataset data}
	\centering
	\begin{tabular}{cccc}
		\hline
		\hline
		Dataset & Image size & Image quantity & Train-Validate-Test\\
		\hline
		\hline
		REFUGE & 512$\times$512 & 1200 & 400-400-400\\
		ACDC & 128$\times$128 & 450 & 100-100-250\\
		CASIA.v4 & 640$\times$480 & 395 & 100-196-99\\
		DTU/Herlev & * & 98 & 20-20-58\\
		\hline
		\hline
	\end{tabular}
\end{table}
\subsubsection{Retinal Fundus Glaucoma Dataset\cite{Refuge}} The Retinal Fundus Glaucoma Challenge (REFUGE) was organized by MICCAI in 2018.
Following the procedure outlined by Liu\cite{DCNNwithSpatialRegularization}, we extract a $512\times 512$ region of interest (ROI) for segmentation. The boundary of the optic cup is more blurred compared to the optic disc, making accurate segmentation of the optic cup region challenging. Considering that the optic cup is predominantly elliptical in shape, we train our model specifically for cup segmentation on this dataset and the learning rate is set to 1e-4 with 100 epochs.
\subsubsection{Automatic Cardiac Diagnosis Challenge Dataset\cite{ACDC}} 

The Automatic Cardiac Diagnosis Challenge (ACDC) was initiated by MICCAI in 2017. This dataset includes MRI scans from 100 patients with expert segmentations.  Given the typically elliptical shape of the left ventricle, we focused on  training and evaluating segmentation performance on this structure. We select data from 10 patients for training, 10 for validation, and 25 for testing. Each patient providing 10 images. On this dataset, the learning rate is set to 5e-5 with 50 epochs.
\subsubsection{CASIA.v4-distance\cite{CASIAv4}} CASIA.v4-distance is a subset of the CASIA.v4 database collected by the Chinese Academy of Sciences' Institute of Automation. In this study, we utilize a subset provided in \cite{CASIAv4label} to train the model for pupil segmentation. The pupil masks were manually annotated by the authors of \cite{CASIAv4label}. We employ 100 of these images for training, utilize 196 images for validation, and reserve the remaining 99 images for testing. On this dataset, the learning rate is set to 1e-4 with 50 epochs.
\subsubsection{DTU/Herlev Dataset\cite{DTUHer}} 
The DTU/Herlev dataset contains 917 single-cell Pap-stained images. 
Given that cell nuclei generally exhibit an elliptical shape, we extract 98 images from this dataset to validate our model. Specifically, 20 images are allocated for training, another 20 for validating, and the remaining 58 for testing. Our model was trained for 50 epochs with a learning rate set at 5e-5.
\subsection{Evaluation Metrics}
\subsubsection{Dice Similarity Coefficient (Dice)} The Dice similarity coefficient between the predicted mask $\bm{P}$ and the ground truth $\bm{G}$ is defined as:
\begin{equation}
	\text{Dice}=\frac{2\left|\bm{P}\cap\bm{G}\right|}{\left|\bm{P}\right|+\left|\bm{G}\right|}
\end{equation}
Dice quantifies the overlap between the predicted and ground truth segmentations within the range $0\le \text{Dice}\le1$, with 1 indicating perfect alignment between the segmentations.
\subsubsection{Boundary Distance (BD)} Boundary Distance assesses the accuracy of object boundaries in segmentation results. Let $e_{\bm{G}}$ and $e_{\bm{P}}$ represent the edge pixels of the ground truth and predicted segmentatioins, respectively. BD is calculated as:
\begin{equation}
	\text{BD}=\sum_{x_i\in e_{\bm{P}}}\frac{D(x_i,e_{\bm{G}})}{\left|e_{\bm{P}}\right|}
\end{equation}
where $D(x_i,e_{\bm{G}})=\min_{x_j\in e_{\bm{G}}}{\|x_i-x_j\|}$, representing the Euclidean distance between pixel $x_i$ on the predicted boundary and its nearest pixel on the ground truth boundary.
\subsubsection{Boundary Distance Standard Deviation (BDSD)} The Boundary Distance Standard Deviation serves as an effective measure to assess the similarity in shape between predicted masks and ground truth values, which is defined as:
\begin{equation}
	\text{BDSD}=\sqrt{\frac{\sum_{x_i\in e_{\bm{P}}}(D(x_i,e_{\bm{G}})-\text{BD})^2}{\left|e_{\bm{P}}\right|}}
\end{equation}
A smaller BDSD value reflects minimal fluctuations in distance between pixels along the segmentation boundary and the corresponding boundary in the ground truth, implying a higher level of shape similarity between the two.
{
\subsection{Training and Inference Time}
During the training process, our proposed shape loss $\mathcal{L}_S$ and ESP module introduces some additional computational overhead. We present the detailed training and inference times of different methods in Table \ref{tab:Trainingtime}.
}
	\begin{table}[!t]
		\centering
		\caption{Computational Efficiency Comparison (Trn T: Training Time per epoch; Inf T: Inference Time per image)}
		\label{tab:Trainingtime}
		\begin{tabular}{c|cc|cc}
			\hline
			\hline
			\multirow{2}{*}{Method} & \multicolumn{2}{c|}{REFUGE} &\multicolumn{2}{c}{ACDC} \\ 
			& Trn T    & Inf T  & Trn T    & Inf T \\ \hline \hline
			SAM-fine      & 1.79min &   80.23ms& 0.34min &   80.50ms\\
			SAM-Sloss     & 2.91min &  80.02ms  & 0.38min &   80.83ms\\
			SAM-ESP      & 4.28min&  178.80ms& 1.01min &   150.16ms \\
			\hline
			\hline
            \multirow{2}{*}{Method} & \multicolumn{2}{c|}{CASIA.v4} &\multicolumn{2}{c}{DTU/Herlev} \\ 
			& Trn T    & Inf T   & Trn T   & Inf T \\ \hline \hline
			SAM-fine      & 0.36min &  80.74ms& 0.10min &   71.73ms\\
			SAM-Sloss     & 1.05min &  81.08ms  & 0.11min &   72.35ms\\
			SAM-ESP      & 1.21min&  166.90ms& 0.44min &   149.50ms \\
			\hline
			\hline
		\end{tabular}
	\end{table}
\subsection{Performances and Analyses}
\begin{table*}[htbp]
	\centering
	\caption{Test Results on Four Datasets}
	\label{tableresult}
	\begin{threeparttable}
		\resizebox{\textwidth}{!}{
			\begin{tabular}{cc|ccc|cc|ccc}
				\hline
				\hline 
				Dataset & Network & Dice$\uparrow$ & BD$\downarrow$ & BDSD$\downarrow$ & Dataset & Network & Dice$\uparrow$ & BD$\downarrow$ & BDSD$\downarrow$ \\
				\hline
				\hline
				\multirow{5}{*}{REFUGE} & SAM & 60.74 & 18.44 & 11.17 &
				\multirow{5}{*}{ACDC} & SAM & 49.03 & 13.32 & 8.12 \\
				\cline{2-5}
				\cline{7-10}
				{} & SAM-fine & 87.62 {$\pm$0.30} & 7.38 {$\pm$0.27} & 4.35 {$\pm$0.11} & {} & SAM-fine & 95.41{$\pm$0.38} & 0.88{$\pm$0.08} & 0.88{$\pm$0.08} \\
				\cline{2-5}
				\cline{7-10}
				{} & SAM-post & 87.76 {$\pm$0.32} & 7.28 {$\pm$0.28} & 4.23 {$\pm$0.14} & {} & SAM-post & 95.50 {$\pm$0.39} & 0.83 {$\pm$0.06} & 0.81 {$\pm$0.05} \\
				\cline{2-5}
				\cline{7-10}
				{} & SAM-Sloss & 88.78 {$\pm$0.15} & 6.67 {$\pm$0.08} & 3.87 {$\pm$0.07} & {} & SAM-Sloss &  {95.51$\pm$0.20} &  {0.86$\pm$0.03} &  {0.90$\pm$0.03} \\
				\cline{2-5}
				\cline{7-10}
				{} & SAM-ESP & \textbf{89.07} {$\pm$0.14} & \textbf{6.36} {$\pm$0.11} & \textbf{3.75} {$\pm$0.13} & {} & SAM-ESP & \textbf{95.85} {$\pm$0.09} & \textbf{0.76} {$\pm$0.004} & \textbf{0.78} {$\pm$0.01} \\
				\hline
				\hline
				\multirow{5}{*}{CASIA.v4} & SAM & 67.29 & 13.85 & 10.85 &
				\multirow{5}{*}{DTU/ Herlev} & SAM & 48.69 & 11.07 & 7.72 \\
				\cline{2-5}
				\cline{7-10}
				{} & SAM-fine & 95.37 {$\pm$0.41} & 1.80 {$\pm$0.10} & 1.23 {$\pm$0.12} & {} & SAM-fine & 95.54 {$\pm$0.37} & 1.77 {$\pm$0.17} & 1.77 {$\pm$0.20} \\
				\cline{2-5}
				\cline{7-10}
				{} & SAM-post & 95.51 {$\pm$0.41} & \textbf{1.74} {$\pm$0.10} & \textbf{1.17} {$\pm$0.10} & {} & SAM-post & 95.80 {$\pm$0.32} & 1.43 {$\pm$0.12} & \textbf{1.28} {$\pm$0.14} \\
				\cline{2-5}
				\cline{7-10}
				{} & SAM-Sloss & 94.12 {$\pm$0.14} & 2.60 {$\pm$0.05} & 1.79 {$\pm$0.06} & {} & SAM-Sloss & 95.92 {$\pm$0.26} & 1.47 {$\pm$0.03} & {1.31} {$\pm$0.03} \\
				\cline{2-5}
				\cline{7-10}
				{} & SAM-ESP & \textbf{95.62} {$\pm$0.18} & {1.74} {$\pm$0.04} & {1.22} {$\pm$0.02} & {} & SAM-ESP & \textbf{96.45} {$\pm$0.16} & \textbf{1.36} {$\pm$0.10} & {1.47} {$\pm$0.18} \\
				\hline
				\hline
			\end{tabular}
		}
		\begin{tablenotes}
			\footnotesize
			\item *SAM denotes the Segment Anything Model without fine-tuning, SAM-fine indicates the fine-tuned Segment Anything Model, and SAM-post refers to the post-processed results of the fine-tuned SAM. SAM-Sloss refers to Segment Anything Model trained with the shape loss.
		\end{tablenotes}
	\end{threeparttable}
\end{table*}	
 Table \ref{tableresult} presents the segmentation performance results on the test sets of various datasets.

The initial SAM model, denoted as SAM, demonstrates suboptimal performance across all datasets, attributed to the inherent dissimilarity between these images and natural scenes. However, fine-tuning, denoted as SAM-fine, significantly improves segmentation accuracy and demonstrates that fine-tuning techniques can adapt models to specific domain.

Furthermore, post-processing the segmentation outputs of SAM, termed SAM-post, yields enhancements, due to the predominant presence of elliptical shapes in our datasets. This refinement step contributes to more accurate delineation of object boundaries, aligning the segmentation results more closely with the expected shapes within the datasets.

Additionally, SAM-Sloss, which incorporates shape constraints into the loss function during model training, showcases improvements in segmentation accuracy.  {While not surpassing SAM-ESP’s performance, SAM-Sloss presents a notable advancement over untuned and finetuned SAM in some instances, suggesting shape constraints in the loss function are viable—albeit at the cost of reduced model stability (consistent with our noise sensitivity findings in Section VI.F).}

\begin{figure*}[tbp]
	\centering
		\subfloat[Images]{
			\begin{minipage}[t]{0.14\textwidth}
				\centering
				\includegraphics[width=1.0in]{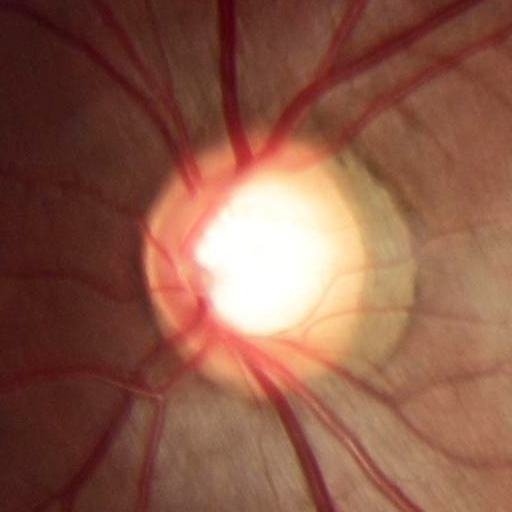}\\
				\vspace{0.2cm}
				\includegraphics[width=1.0in]{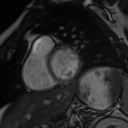}\\
				\vspace{0.2cm}
				\includegraphics[width=1.0in]{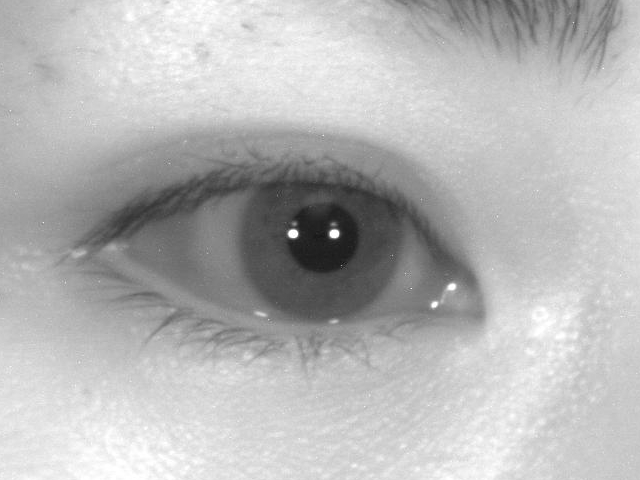}\\
				\vspace{0.2cm}
				\includegraphics[width=1.0in]{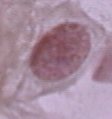}\\
				\vspace{0.05cm}
			\end{minipage}%
		}%
		\hspace{0.1cm}
		\subfloat[{SAM}]{
			\begin{minipage}[t]{0.14\linewidth}
				\centering
				\includegraphics[width=1.0in]{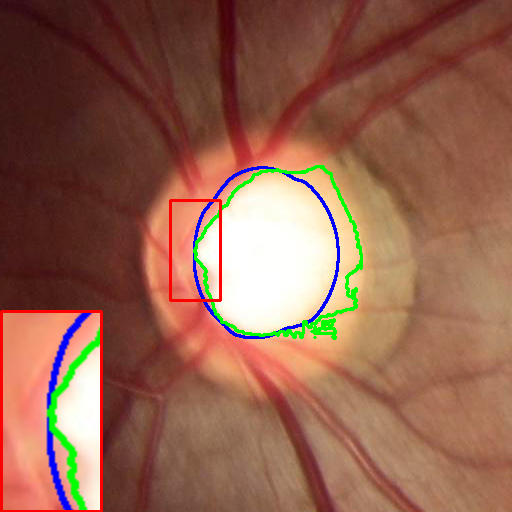}\\
				\vspace{0.2cm}
				\includegraphics[width=1.0in]{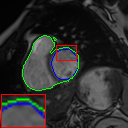}\\
				\vspace{0.2cm}
				\includegraphics[width=1.0in]{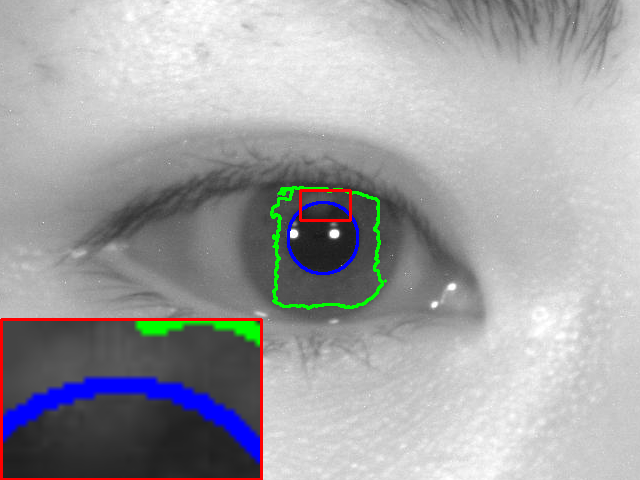}\\
				\vspace{0.2cm}
				\includegraphics[width=1.0in]{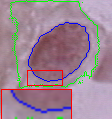}\\
				\vspace{0.05cm}
			\end{minipage}%
		}%
		\hspace{0.1cm}
		\subfloat[{SAM-fine}]{
			\begin{minipage}[t]{0.14\linewidth}
				\centering
				\includegraphics[width=1.0in]{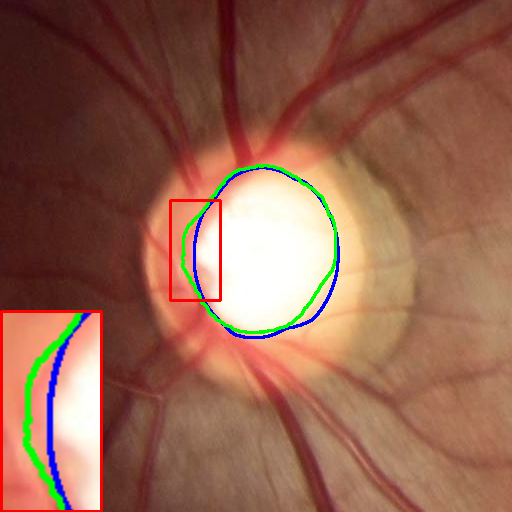}\\
				\vspace{0.2cm}
				\includegraphics[width=1.0in]{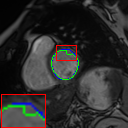}\\
				\vspace{0.2cm}
				\includegraphics[width=1.0in]{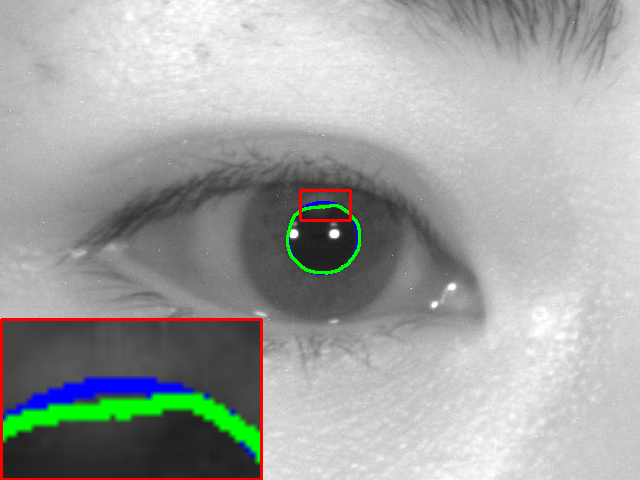}\\
				\vspace{0.2cm}
				\includegraphics[width=1.0in]{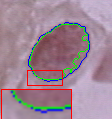}\\
				\vspace{0.05cm}
			\end{minipage}%
		}%
		\hspace{0.1cm}
		\subfloat[SAM-post]{
			\begin{minipage}[t]{0.14\linewidth}
				\centering
				\includegraphics[width=1.0in]{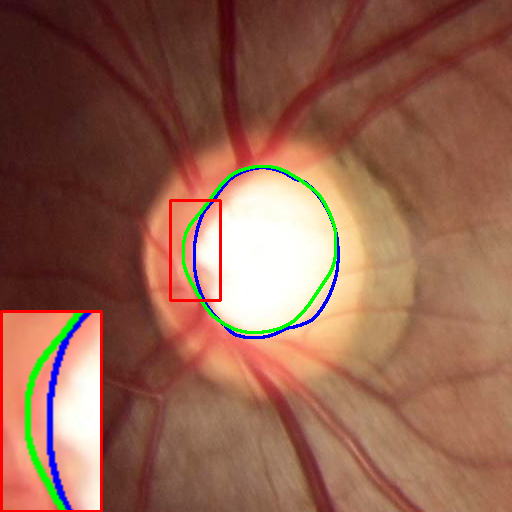}\\
				\vspace{0.2cm}
				\includegraphics[width=1.0in]{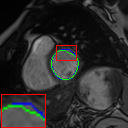}\\
				\vspace{0.2cm}
				\includegraphics[width=1.0in]{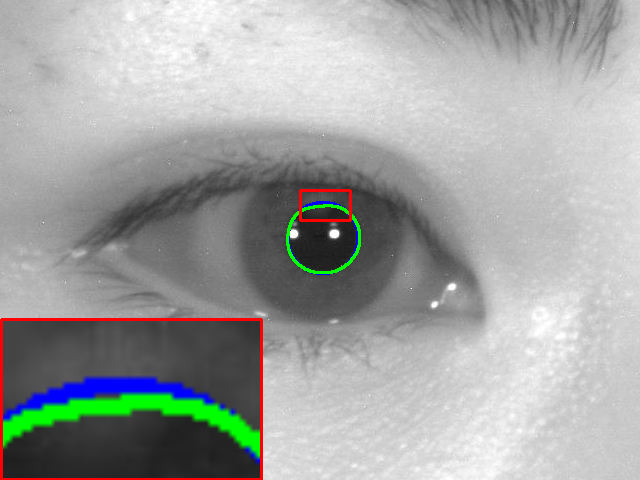}\\
				\vspace{0.2cm}
				\includegraphics[width=1.0in]{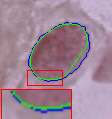}\\
				\vspace{0.05cm}
			\end{minipage}%
		}%
		\hspace{0.1cm}
		\subfloat[SAM-Sloss]{
			\begin{minipage}[t]{0.14\linewidth}
				\centering
				\includegraphics[width=1.0in]{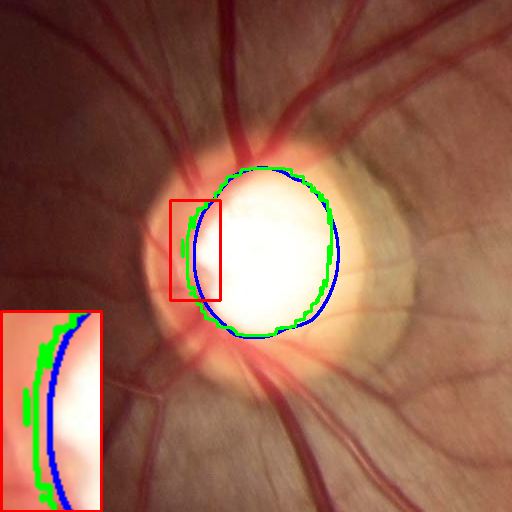}\\
				\vspace{0.2cm}
				\includegraphics[width=1.0in]{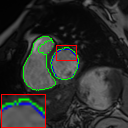}\\
				\vspace{0.2cm}
				\includegraphics[width=1.0in]{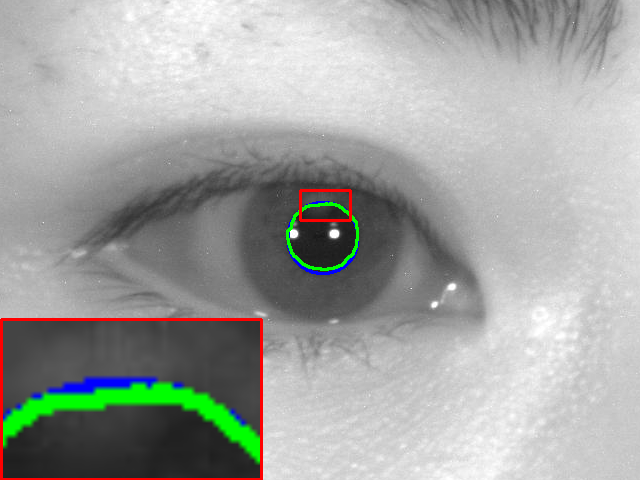}\\
				\vspace{0.2cm}
				\includegraphics[width=1.0in]{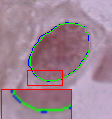}\\
				\vspace{0.05cm}
			\end{minipage}%
		}%
		\hspace{0.1cm}
		\subfloat[SAM-ESP]{
			\begin{minipage}[t]{0.14\linewidth}
				\centering
				\includegraphics[width=1.0in]{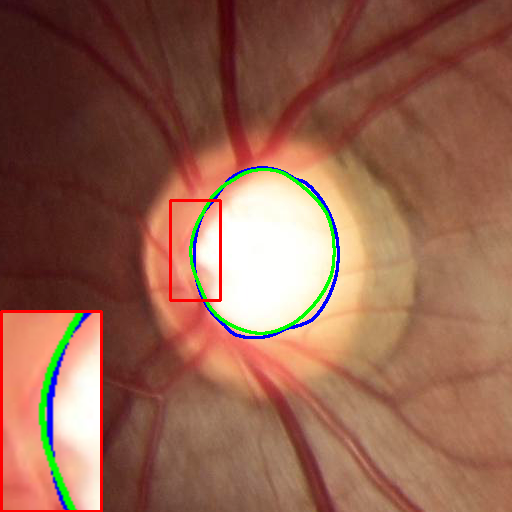}\\
				\vspace{0.2cm}
				\includegraphics[width=1.0in]{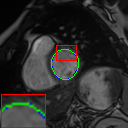}\\
				\vspace{0.2cm}
				\includegraphics[width=1.0in]{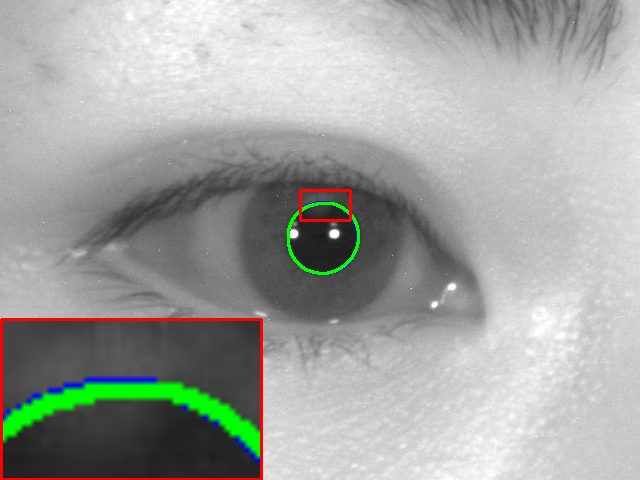}\\
				\vspace{0.2cm}
				\includegraphics[width=1.0in]{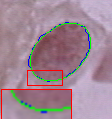}\\
				\vspace{0.05cm}
			\end{minipage}%
	}
	\centering
	\caption{Visual effects of the segmentation results on four datasets. The blue line in the figure represents the segmentation contour of the ground truth, while the green line depicts the segmentation contours generated by various models. The red boxes indicate locally magnified views.}
	\label{testresult}
\end{figure*}

Among all evaluated models, SAM-ESP emerges as the top performer, showcasing high segmentation accuracy across all metrics, particularly in terms of the Dice coefficient. SAM-ESP represents a novel approach where shape constraints are seamlessly integrated into the network architecture during model design. While SAM-Sloss and SAM-post  {exhibits} commendable enhancements, SAM-ESP's superior performance underscores its efficacy in preserving the elliptical shapes present in images. 
For a more intuitive understanding, visual comparisons of the experimental outcomes are presented in~\figurename\ref{testresult}.

\subsection{Generalization Ability for Noise}
\begin{figure}[htbp]
    \centering
    \begin{minipage}[b]{0.48\linewidth}
        \subfloat[\small {Gaussian} noise]{%
            \includegraphics[width=\linewidth]{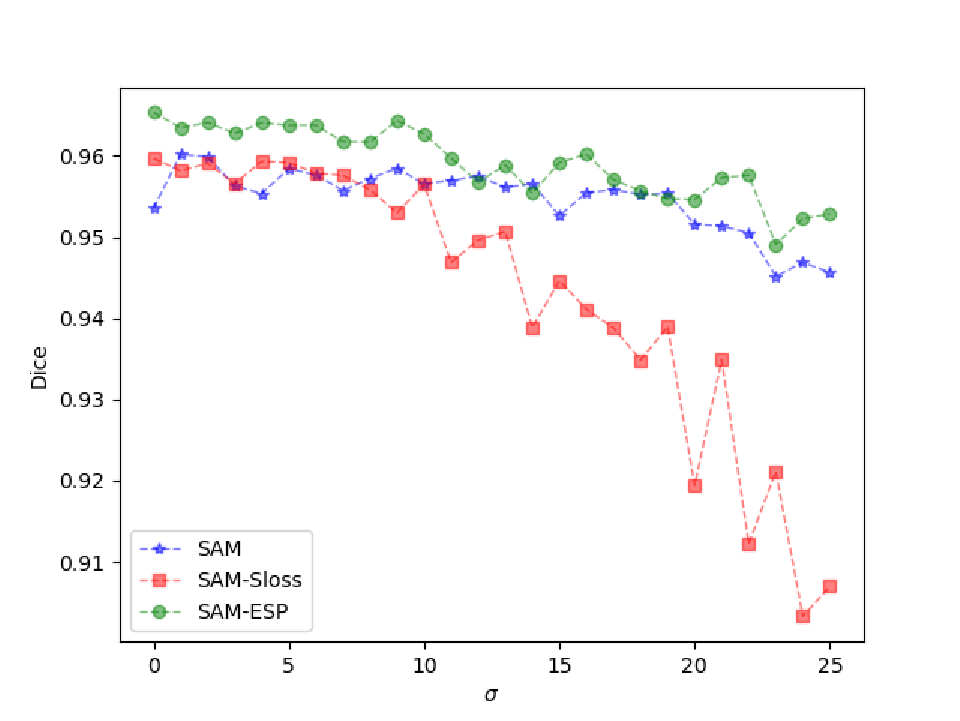}%
            \label{guassian}%
        }%
    \end{minipage}
    \hspace{0.03cm}
    \begin{minipage}[b]{0.48\linewidth}
        \subfloat[\small {salt-and-pepper noise}]{%
            \includegraphics[width=\linewidth]{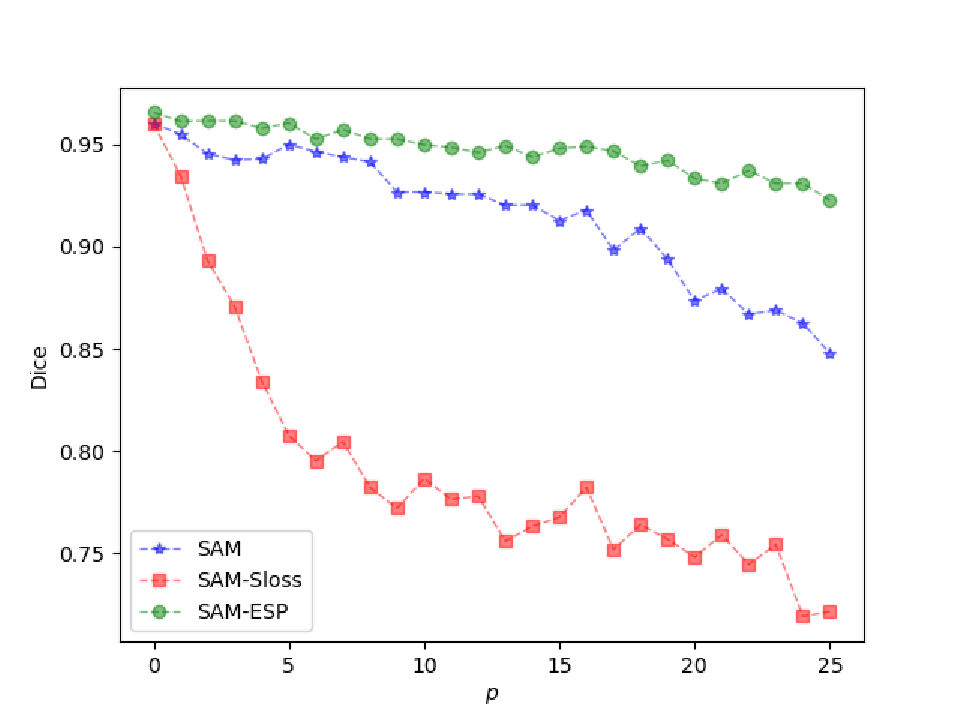}%
            \label{saltpepper}%
        }
    \end{minipage}
    \caption{The generalization ability for different noise.}
    \label{dice with noised}
\end{figure}

To evaluate the generalization ability of ESP module under varying levels of noise, we introduced Gaussian noise with standard deviations ($\sigma$) ranging from 0 to 25 separately onto the DTU/Herlev test set. Additionally, to assess the model's robustness against different types of noise, we introduced salt-and-pepper noise ($p$) ranging from 0 to 25\% of image pixels on the same test set.  \figurename\ref{guassian} and \figurename\ref{saltpepper} respectively illustrate the Dice scores of three variants of the SAM model: SAM-fine, SAM trained with shape loss (SAM-Sloss), and SAM-ESP, under varying degrees of Gaussian and salt-and-pepper noise influences.

\begin{figure*}[htbp]
	\centering
	\subfloat[Image]{
		\begin{minipage}[t]{0.16\linewidth}
		\centering
		\includegraphics[width=\linewidth]{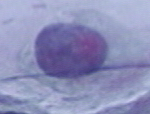}\\
		\vspace{0.2cm}
		\includegraphics[width=\linewidth]{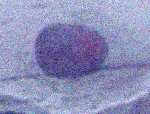}\\
		\vspace{0.2cm}
		\includegraphics[width=\linewidth]{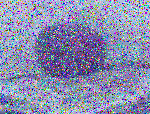}
	\end{minipage}
	}
	\hspace{0.1cm}
	\subfloat[Ground Truth]{
		\begin{minipage}[t]{0.16\linewidth}
		\centering
		\includegraphics[width=\linewidth]{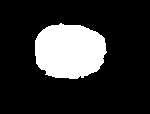}\\
		\vspace{0.2cm}
		\includegraphics[width=\linewidth]{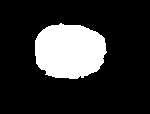}\\
		\vspace{0.2cm}
		\includegraphics[width=\linewidth]{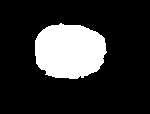}
	\end{minipage}
	}
	\hspace{0.1cm}
	\subfloat[SAM-fine]{
		\begin{minipage}[t]{0.16\linewidth}
		\centering
		\includegraphics[width=\linewidth]{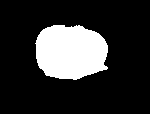}\\
		\vspace{0.2cm}
		\includegraphics[width=\linewidth]{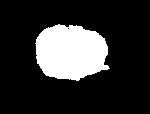}\\
		\vspace{0.2cm}
		\includegraphics[width=\linewidth]{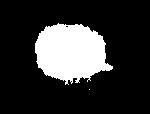}
		\end{minipage}
	}
	\hspace{0.1cm}
	\subfloat[SAM-Sloss]{
		\begin{minipage}[t]{0.16\linewidth}
		\centering
		\includegraphics[width=\linewidth]{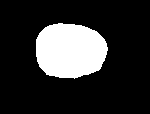}\\
		\vspace{0.2cm}
		\includegraphics[width=\linewidth]{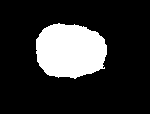}\\
		\vspace{0.2cm}
		\includegraphics[width=\linewidth]{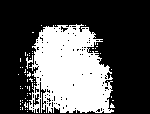}
		\end{minipage}
	}
	\hspace{0.1cm}
	\subfloat[SAM-ESP]{
		\begin{minipage}[t]{0.16\linewidth}
		\centering
		\includegraphics[width=\linewidth]{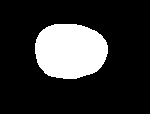}\\
		\vspace{0.2cm}
		\includegraphics[width=\linewidth]{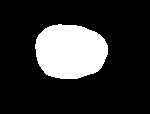}\\
		\vspace{0.2cm}
		\includegraphics[width=\linewidth]{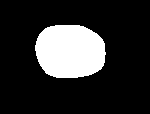}
		\end{minipage}
	}
	\caption{Visualization of model segmentation results under noise influence.  The first row displays segmentation results without noise. The second row displays segmentation results under Gaussian noise with a standard deviation of 20, while the third row shows segmentation results under 20\% salt-and-pepper noise.}
	\label{Dnoise}      
\end{figure*}

\begin{table}[htbp]
	\centering
	\caption{Test Results Under Noise Influence on DTU/Herlev Datasets}
	\label{noiseresult}
	\begin{threeparttable}
		\begin{tabular}{cc|ccc}
			\hline
			\hline 
		  Noise & Network & Dice$\uparrow$ & BD$\downarrow$ & BDSD$\downarrow$ \\
			\hline
			\hline
			 \multirow{3}{*}{GS} & SAM-fine & 95.12 {$\pm$0.54} & 2.07 {$\pm$0.17} & 2.20 {$\pm$0.17} \\
			  {} & SAM-Sloss & {93.13} {$\pm$0.18} & {2.56} {$\pm$0.06} & 2.14 {$\pm$0.15} \\
			 {} & SAM-ESP & \textbf{95.31} {$\pm$0.20} & \textbf{1.93} {$\pm$0.15} & \textbf{2.03} {$\pm$0.27} \\
			\hline
			 \multirow{3}{*}{SP} & SAM-fine & 89.97 {$\pm$0.52} & 4.71 {$\pm$0.23} & 4.32 {$\pm$0.19} \\
			  {} & SAM-Sloss & 76.71 {$\pm$0.41} & 8.52 {$\pm$0.22} & 6.03 {$\pm$0.15} \\
              {} & SAM-ESP & \textbf{93.75} {$\pm$1.07} & \textbf{2.51} {$\pm$0.32} & \textbf{2.34} {$\pm$0.34} \\
			\hline
			\hline
		\end{tabular}
	\begin{tablenotes}
		\footnotesize
		\item *'GS' denotes the  {Gaussian} noise while 'SP' denotes the salt-and-pepper noise.
	\end{tablenotes}
	\end{threeparttable}
\end{table}	
To facilitate comprehensive comparison across various metrics, we conducted a re-evaluation of the three models under specific conditions: a standard deviation of Gaussian noise set to 20 and 20\% pixel contamination due to salt-and-pepper noise. The corresponding results are presented in \figurename\ref{Dnoise} and Table \ref{noiseresult}.
From \figurename\ref{Dnoise}, it can be observed that the addition of noise leads to less smooth boundaries in the segmentation results of SAM-fine. Particularly under the influence of salt-and-pepper noise, SAM-Sloss fails to correctly segment cell nuclei, whereas SAM-ESP maintains satisfactory segmentation results. Furthermore, from \figurename\ref{dice with noised}, it is evident that as the intensity of noise increases, the segmentation accuracy of all three models begins to decline. Notably, SAM-Sloss exhibits a significant decrease in segmentation accuracy compared to the other two models, indicating pronounced sensitivity to noise. This sensitivity may stem from the shape prior in the loss function, which directs the model's focus more towards the current task during training at the expense of some generalization ability. Additionally, \figurename\ref{dice with noised} demonstrates that SAM-ESP consistently maintains higher segmentation accuracy under different noise conditions, with smaller declines in accuracy, particularly evident under salt-and-pepper noise interference. This stable performance underscores the superior robustness of SAM-ESP to noise interference.

 {
\subsection{Generalization Analysis on External Datasets}
In this section, we evaluate the generalization capability of our models trained on the REFUGE dataset by testing them on two external datasets with distinct data distributions: RIM-ONE DL \cite{RIMONEDL} and BinRushed (RIGA)\cite{RIGA,RIGA_mask}. The RIM-ONE DL dataset comprises 118 test images with optic cup masks from healthy subjects, while BinRushed contains 195 images with annotated optic cup and disc boundaries. All BinRushed images were center-cropped to 512×512 for input standardization.}
\begin{table}[htbp]
	\centering
	\caption{Test Results on External Datasets }
	\label{zero-shot}
    \resizebox{\linewidth}{!}{%
		\begin{tabular}{cc|ccc}
			\hline
			\hline 
		 Dataset  & Network & Dice$\uparrow$ & BD$\downarrow$ & BDSD$\downarrow$ \\
			\hline
			\hline
			\multirow{3}{*}{RIM-ONE DL} & SAM-fine & 83.06$\pm$0.52 & 9.22$\pm$0.26 & 5.19$\pm$0.09 \\
			 {} & SAM-Sloss & {82.33$\pm$0.79} & {9.71$\pm$0.45} & 5.46$\pm$0.28 \\
			{} & SAM-ESP & \textbf{83.68}$\pm$0.41 & \textbf{8.78}$\pm$0.14 & \textbf{4.96}$\pm$0.05 \\
			\hline
            \hline
			\multirow{3}{*}{BinRushed} & SAM-fine & 79.55$\pm$2.52 & 8.42$\pm$0.78 & 4.72$\pm$0.27 \\
			 {}  & SAM-Sloss & 79.08$\pm$2.40 & 8.77$\pm$0.53 & 4.88$\pm$0.20 \\
             {}  & SAM-ESP & \textbf{80.51}$\pm$2.84 & \textbf{7.45}$\pm$0.50 & \textbf{4.17}$\pm$0.21 \\
			\hline
			\hline
		\end{tabular}}
\end{table}	

\begin{figure}[htbp]
	\centering
        \captionsetup[subfloat]{labelformat=empty}
		\subfloat[\small Image]{
			\begin{minipage}[b]{0.22\linewidth}
				\includegraphics[width=1\linewidth]{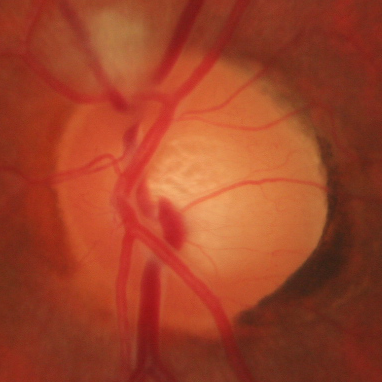}\vspace{2pt}
                \includegraphics[width=1\linewidth]{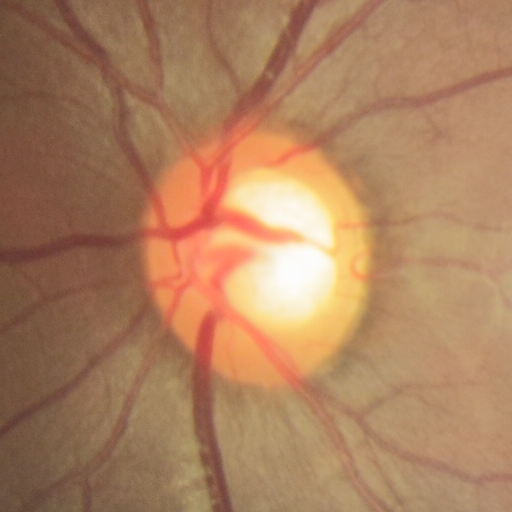}
			\end{minipage}
            \vspace{-4pt}
	}
		\subfloat[\small SAM-fine]{
			\begin{minipage}[b]{0.22\linewidth}
				\includegraphics[width=1\linewidth]{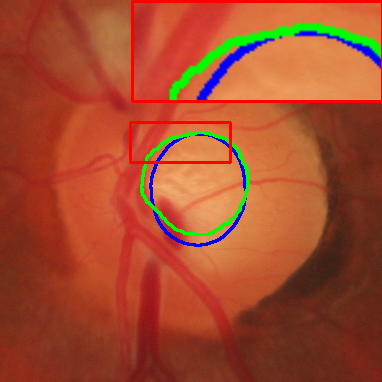}\vspace{2pt}
				\includegraphics[width=1\linewidth]{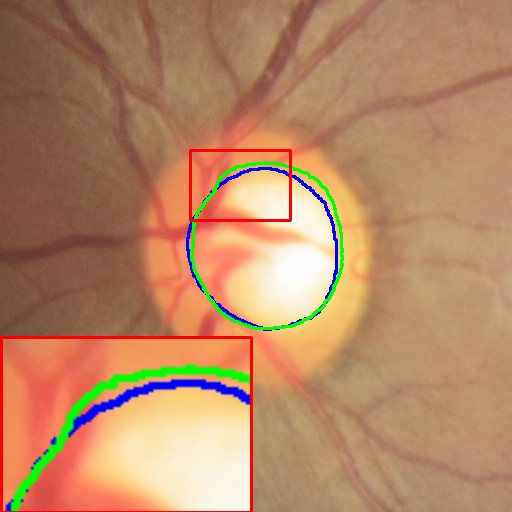}
			\end{minipage}
            \vspace{-4pt}
	}
		\subfloat[\small SAM-Sloss]{
			\begin{minipage}[b]{0.22\linewidth}
				\includegraphics[width=1\linewidth]{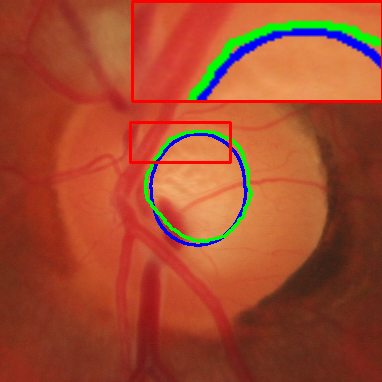}\vspace{2pt}
				\includegraphics[width=1\linewidth]{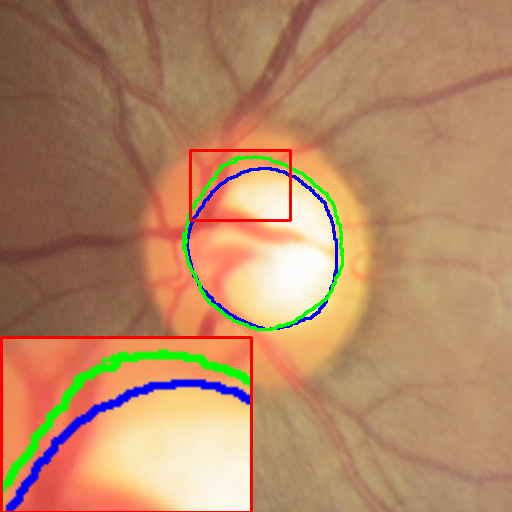}
			\end{minipage}
            \vspace{-4pt}
		}
		\subfloat[\small SAM-ESP]{
			\begin{minipage}[b]{0.22\linewidth}
				\includegraphics[width=1\linewidth]{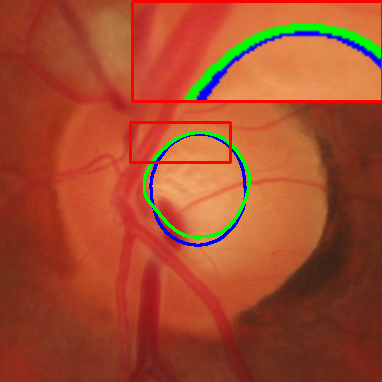}\vspace{2pt}
				\includegraphics[width=1\linewidth]{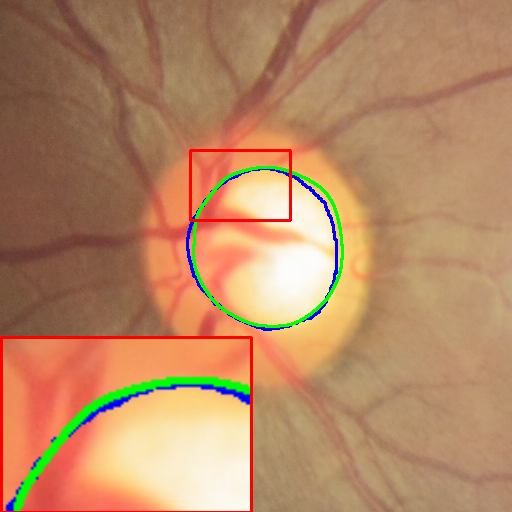}
			\end{minipage}
            \vspace{-4pt}
		}
        \vspace{-0.08cm}
		\caption{Visualization of segmentation results on external datasets. Blue: ground truth contours; green: model predictions. Red boxes highlight local details. The first row shows results on the RIM-ONE DL dataset, while the second row displays results on the BinRushed dataset.}
		\label{fig:External_result}
	\end{figure}

 {As demonstrated in Table \ref{zero-shot}, our SAM-ESP achieves superior performance on both external datasets (RIM-ONE DL and BinRushed) compared to the fine-tuned SAM baseline. These results indicate that SAM-ESP exhibits strong cross-dataset generalization capability under varying data distributions. Moreover, the experimental results reveal that solely modifying the loss function degrades the model's generalization ability. The qualitative results in \figurename\ref{fig:External_result} confirm the advantages of our module in segmentation boundary accuracy.}

\begin{figure}[htbp]
		\centering
        \captionsetup[subfloat]{labelformat=empty}
	       \subfloat[\small SAMoph]{
			\begin{minipage}[b]{0.17\linewidth}
				\includegraphics[width=\linewidth]{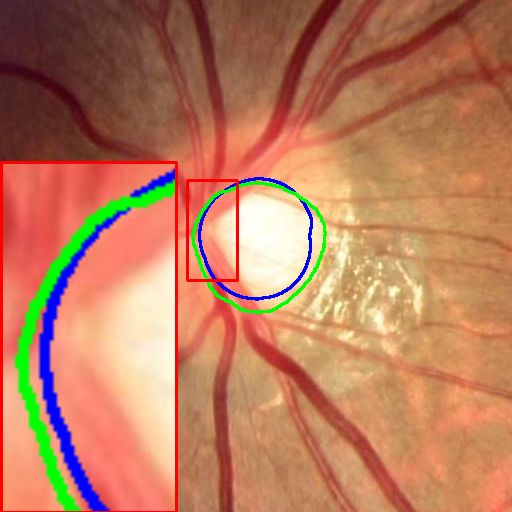}\\[2pt]
                \includegraphics[width=\linewidth]{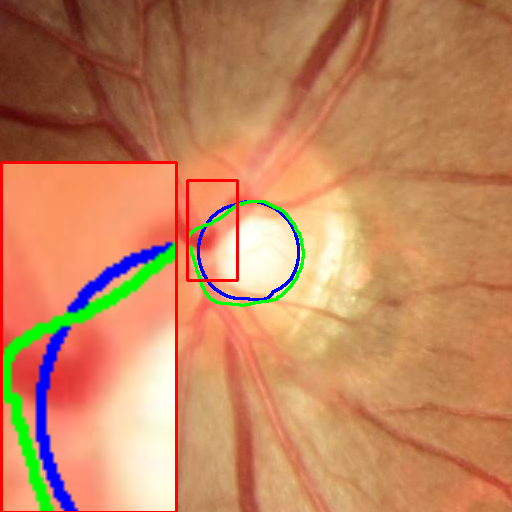}\\[2pt]
                \includegraphics[width=\linewidth]{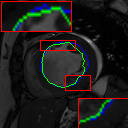}\\[2pt]
                \includegraphics[width=\linewidth]{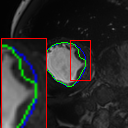}
			\end{minipage}
            \vspace{-4pt}
	}
		\subfloat[\small SAM-$\mathcal{L}_{cs}$]{
			\begin{minipage}[b]{0.17\linewidth}
				\includegraphics[width=\linewidth]{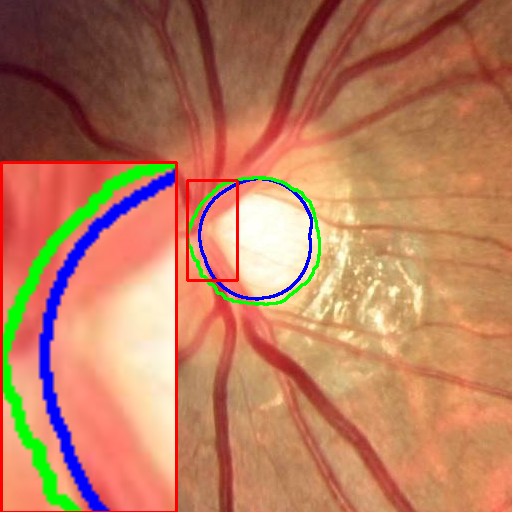}\\[2pt]
				\includegraphics[width=\linewidth]{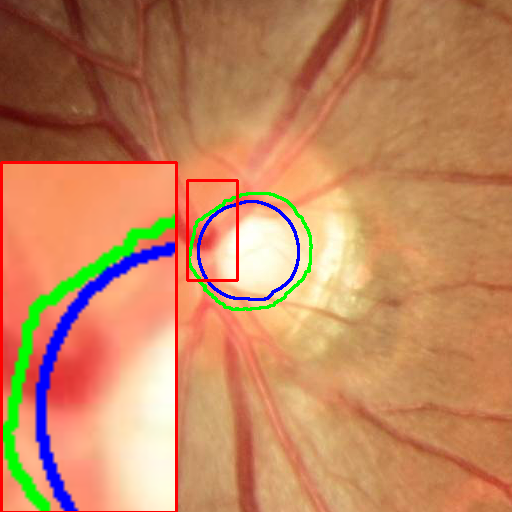}\\[2pt]
                \includegraphics[width=\linewidth]{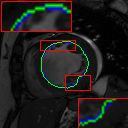}\\[2pt]
                \includegraphics[width=\linewidth]{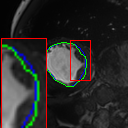}
			\end{minipage}
            \vspace{-4pt}
	}
		\subfloat[\small SAM-$\mathcal{L}_{efe}$]{
			\begin{minipage}[b]{0.17\linewidth}
				\includegraphics[width=\linewidth]{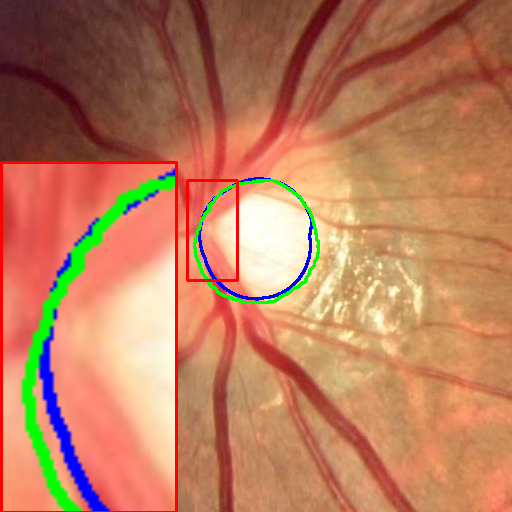}\\[2pt]
				\includegraphics[width=\linewidth]{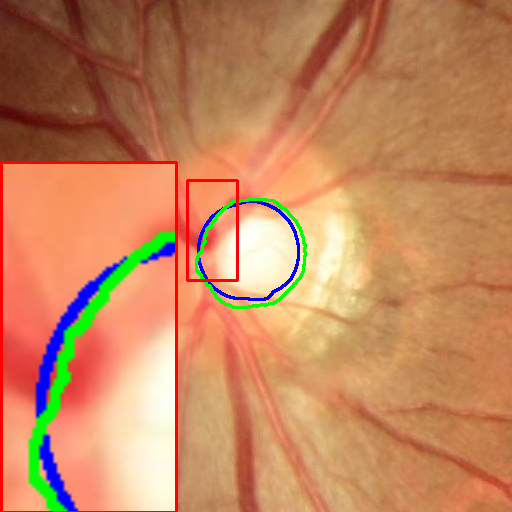}\\[2pt]
                \includegraphics[width=\linewidth]{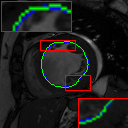}\\[2pt]
                \includegraphics[width=\linewidth]{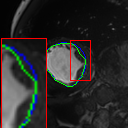}
			\end{minipage}
            \vspace{-4pt}
		}
		\subfloat[\small SAM-Sloss]{
			\begin{minipage}[b]{0.17\linewidth}
				\includegraphics[width=\linewidth]{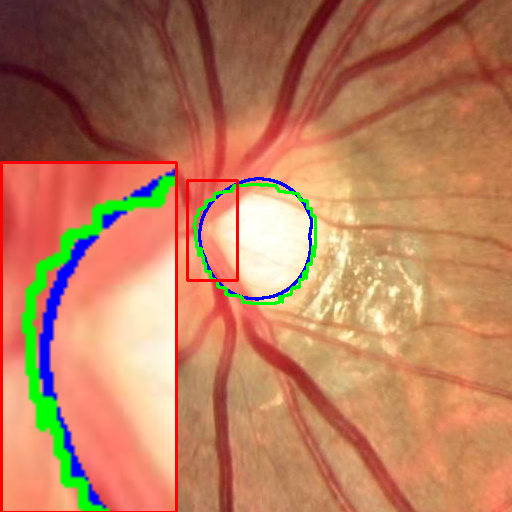}\\[2pt]
				\includegraphics[width=\linewidth]{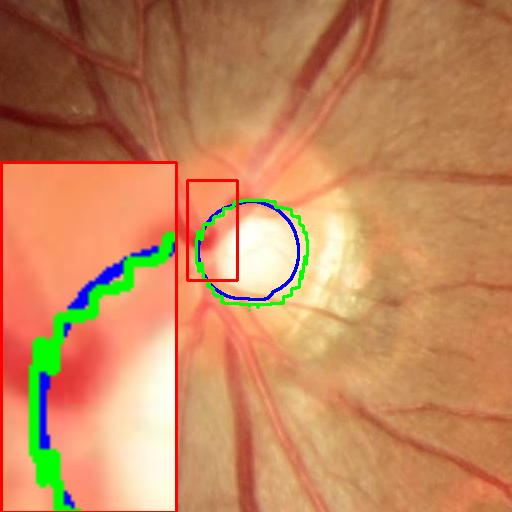}\\[2pt]
                \includegraphics[width=\linewidth]{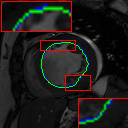}\\[2pt]
                \includegraphics[width=\linewidth]{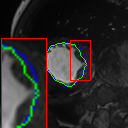}
			\end{minipage}
            \vspace{-4pt}
		}
        \subfloat[\small SAM-ESP]{
			\begin{minipage}[b]{0.17\linewidth}
				\includegraphics[width=\linewidth]{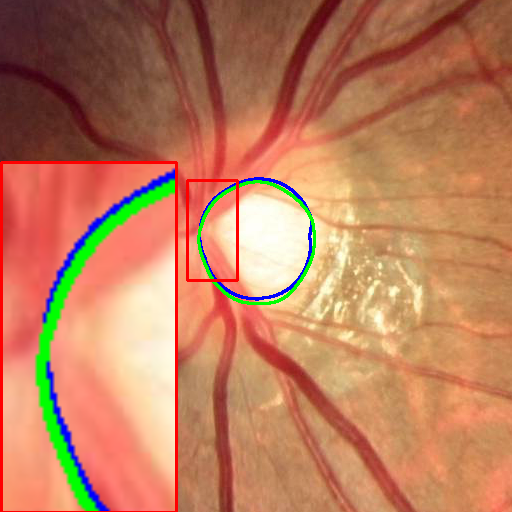}\\[2pt]
				\includegraphics[width=\linewidth]{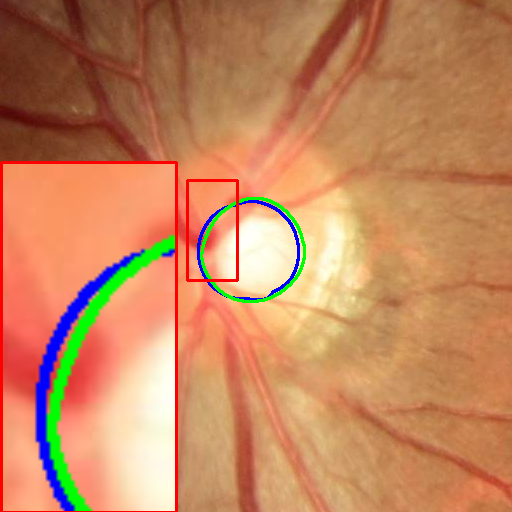}\\[2pt]
                \includegraphics[width=\linewidth]{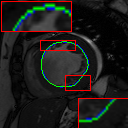}\\[2pt]
                \includegraphics[width=\linewidth]{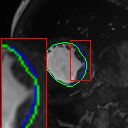}
			\end{minipage}
            \vspace{-4pt}
		}
    \vspace{2pt}
		\caption{Segmentation results comparison. Blue: ground truth contours; green: model predictions. Red boxes highlight local details.}
		\label{fig:related_work}
	\end{figure}

 {
\subsection{Comparative Analysis}
In this experimental section, we systematically compare the proposed method with three related state-of-the-art approaches on the REFUGE and ACDC datasets: Convex shape prior loss $\mathcal{L}_{cs}$~\cite{convexshapeloss} enforces convexity constraints by penalizing cases where line segments between any two points within the segmented region fall outside the area; Ellipse fit error $\mathcal{L}_{efe}$~\cite{CNNpuilsegmentation} constrains geometric shapes by minimizing the distance between predicted segmentation boundaries and their best-fit ellipses; while Learnable Ophthalmology SAM~\cite{SAMoph} incorporates learnable prompt layers into SAM's encoder to enhance medical image adaptation. To ensure fair comparison, all methods follow the identical training-testing protocol: 100 epochs (LR=1e-4) for REFUGE and 50 epochs (LR=5e-5) for ACDC, with three independent runs per method. Test results are reported in Table \ref{tab:comparative_experiments}. Some examples are also shown in \figurename\ref{fig:related_work}. From the experimental results, it can be seen that our SAM-ESP model better preserves the elliptical contour of the segmented region.
}
\begin{table}[htbp]
		\centering
		\caption{Comparative Experiments with the Most Related Methods}
		\label{tab:comparative_experiments}
		\begin{tabular}{c|ccc}
			\hline
			\hline
			\multirow{2}{*}{Method} & \multicolumn{3}{c}{REFUGE} \\ 
			& {Dice}$\uparrow$    & {BD}$\downarrow$    & {BDSD}$\downarrow$   \\ \hline \hline
                SAMoph\cite{SAMoph} & 84.54$\pm$0.10 &9.49$\pm$0.14&  5.60$\pm$0.09  \\
			SAM-$\mathcal{L}_{cs}$\cite{convexshapeloss}    & 87.00$\pm$0.54 &  7.75$\pm$0.37 &  4.47$\pm$0.18  \\
	SAM-$\mathcal{L}_{efe}$\cite{CNNpuilsegmentation}      & 88.36$\pm$0.57&  6.82$\pm$0.33 & 4.05$\pm$0.21    \\
                SAM-Sloss & 88.78{$\pm$0.15} & 6.67{$\pm$0.08} & 3.87{$\pm$0.07} \\
                SAM-ESP & \textbf{89.07}{$\pm$0.14} & \textbf{6.36}{$\pm$0.11} & \textbf{3.75}{$\pm$0.13} \\
			\hline
			\hline
            \multirow{2}{*}{Method} & \multicolumn{3}{c}{ACDC} \\ 
			& {Dice}$\uparrow$    & {BD}$\downarrow$    & {BDSD}$\downarrow$   \\ \hline \hline
                SAMoph\cite{SAMoph} & 92.98$\pm$0.48 &1.28$\pm$0.18&  0.93$\pm$0.03  \\
			SAM-$\mathcal{L}_{cs}$\cite{convexshapeloss}    &95.20$\pm$0.08 &  0.87$\pm$0.006 &  0.81$\pm$0.02  \\
	SAM-$\mathcal{L}_{efe}$\cite{CNNpuilsegmentation}      & 95.63$\pm$0.07&  0.87$\pm$0.03 & 0.89$\pm$0.03    \\
                SAM-Sloss &95.51 {$\pm$0.20} & 0.86 {$\pm$0.03} & 0.90 {$\pm$0.03} \\
                SAM-ESP & \textbf{95.85} {$\pm$0.09} & \textbf{0.76} {$\pm$0.004} & \textbf{0.78} {$\pm$0.01} \\
			\hline
			\hline
		\end{tabular}
	\end{table}

 {
\subsection{Ablation Study for ESP Module}
In this section, we investigate the impact of different depths of the ESP module on segmentation performance. Specifically, we vary the number of algorithm iterations $T$, starting from $T=75$ and increasing by 25 at each step. Experiments on the REFUGE dataset (Table \ref{tab:number_of_layer}) indicate that the best performance is achieved when $T=100$. The ESP module is derived from our variational segmentation model, where each layer corresponds to one iteration of the algorithm. Thus, a sufficient number of iterations is required to achieve optimal results. However, during network training, computational complexity must also be considered, as excessively large 
$T$ values may degrade the training process.
	\begin{table}[htbp]
		\centering
		\caption{Ablation Study on the Depths of ESP Module}
		\label{tab:number_of_layer}
		\begin{tabular}{c|ccc}
			\hline
			\hline
			\multirow{2}{*}{ESP module} & \multicolumn{3}{c}{REFUGE} \\ 
			& {Dice}$\uparrow$    & {BD}$\downarrow$    & {BDSD}$\downarrow$   \\ \hline \hline
			$T=75$      & 88.34$\pm$0.07 &   6.79$\pm$0.10&  3.99$\pm$0.07  \\
			$T=100$     &\textbf{89.07}{$\pm$0.14} & \textbf{6.36}{$\pm$0.11} & \textbf{3.75}{$\pm$0.13}  \\
			$T=125$      & 88.05$\pm$0.51&  7.16$\pm$0.34 &  4.20$\pm$0.15    \\
                $T=150$      & 88.07$\pm$0.38 &   7.00$\pm$0.21&  4.13$\pm$0.09 \\
			\hline
			\hline
		\end{tabular}
	\end{table}
}

\section{Conclusion}
\label{conclusion}
We have presented a novel segmentation algorithm that integrates an elliptical shape prior into the variational model through the constraint of the contour field. Our algorithm is rooted in the primal-dual theory,  {showing efficiency in practical experiments}. The crucial sub-problem in our algorithm  {admits a softmax closed-form solution}, which allows for the integration of the elliptical shape, spatial regularization, and other priors in the model. This introduces a fresh perspective on incorporating elliptical priors into deep learning-based SAM image segmentation methods. Comparative results on different types of images showcase the robust performance of the proposed model in accurately segmenting elliptical regions.

Our method can be extended to any other deep neural network approaches for image segmentation.  {If the image segmentation dataset consists of elliptical shapes}, our method can be used to construct an image segmentation deep network structure, ensuring that the network's output adheres to the elliptical shape prior. 

 It is worth noting that our method has certain limitations. If the target objects are not all elliptical shapes, using such a strong prior will instead cause all segmented objects to become approximately elliptical. Therefore, exploring a more general prior shape will help broaden the application scope of the proposed method. This will be an important direction for our future research.

\section*{Acknowledgments}
Jun Liu was supported by the 
National Key Research and Development Program
of China (No. 2023YFC3008505), Jun Liu and Faqiang Wang were supported by the National Natural Science Foundation of China (No.12371527, No. 42293272) and the Beijing Natural Science Foundation (No. 1232011). Li Cui was supported by the National Natural Science Foundation of China (No. 12171043). Faqiang Wang was supported by the National Natural Science Foundation of China (No. 12101058). This research was also supported by Super Computing Center of Beijing Normal University, user name is liujun.

{\appendices
\section{Proof of Proposition \ref{proposition1}}
\label{appendix_A}
	\begin{IEEEproof}
			If the contours of $u(x,y)$ are ellipses, then $\exists~\phi(t), \psi(t)$ defined by (\ref{ell-equ}) such that
		$u(\phi(t),\psi(t))=c$. Here $c$ is a constant. Since $u$ is $\mathcal{C}^1$, differentiate both sides of the above equation with respect to $t$, we get $u_x\phi^{'}(t)+u_y\psi^{'}(t)=0$, i.e. $\langle\nabla u(x,y), \bm{T}_{\Lambda}(x,y)\rangle=0.$\\
		\indent Conversely, we prove by contradiction. Assume that when the conditions are satisfied, there is a contour of $u$ is not ellipse. Denoting this contour as $u(x(t),y(t))=c$, then we have  $u_xx^{'}(t)+u_yy^{'}(t)=0$. 
		From the given conditions, we have the 2D vectors $(x^{'}(t),y^{'}(t))$ and $(\phi^{'}(t),\psi^{'}(t))$ are both orthogonal to
		$\nabla u$ , thus $(x^{'}(t),y^{'}(t))$ and $(\phi^{'}(t),\psi^{'}(t))$ are parallel. This implies that $(\phi(t),\psi(t))$ is not an ellipse. Contradiction! Therefore, the assumption is not valid, which completes the proof.
	\end{IEEEproof}}
\section{Calculation of $\bm{u}^{t+1}$}
\label{appendix_B}
   {
    The subproblem of $\bm{u}$ is:
    \begin{equation*}
	\bm{u}^{t+1}=\underset{\bm{u}\in\mathbb{U}}{\arg\min}\left\{
	\begin{aligned}
	&\langle-\bm{o}+\bm{p}^t,\bm{u}\rangle+\varepsilon\langle\bm{u},\ln(\bm{u})\rangle\\
	&\quad+\langle \text{div}(q^{t+1}\bm{T}^t),u_{i}\rangle
		\end{aligned}
	\right\}.		
\end{equation*}
    For the simplicity of notation, let $\delta_{\hat{i},i}$ be the delta function such that $\delta_{i,i}=1$ and $\delta_{\hat{i},i}=0, \hat{i}\neq i$ otherwise. We define $r_{\hat{i}}:=-o_{\hat{i}}+p_{\hat{i}}^t+\delta_{\hat{i},i}\text{div}(q^{t+1}\bm{T}^t)$ then the subproblem can be formulated as 
    \begin{equation*}
    \begin{aligned}
    &\underset{\bm{u}\ge 0}{\min}\left\{\sum_{\hat{i}=1}^I\int_\Omega{r_{\hat{i}}(x)u_{\hat{i}}(x)}\mathrm{d}x+\int_{\Omega}\varepsilon u_{\hat{i}}(x)\ln{u_{\hat{i}}(x)}\mathrm{d}x\right\}\\
    &s.t. \sum_{\hat{i}=1}^Iu_{\hat{i}}(x)=1,\quad \forall x\in\Omega.
    \end{aligned}
\end{equation*}
    By introducing Lagrangian multiplier $v$ associated to the constraint $\sum\limits_{\hat{i}=1}^Iu_{\hat{i}}(x)=1,\forall x\in\Omega$, we have the related Lagrangian function:
    \begin{equation*}
    \begin{aligned}
        \mathcal{L}(\bm{u},v)=&\sum_{\hat{i}=1}^I\int_\Omega{r_{\hat{i}}(x)u_{\hat{i}}(x)}\mathrm{d}x+\sum_{\hat{i}=1}^I\int_{\Omega}\varepsilon u_{\hat{i}}(x)\ln{u_{\hat{i}}(x)}\mathrm{d}x\\
        &+\int_{\Omega}{v(x)}(1-\sum_{\hat{i}=1}^Iu_{\hat{i}}(x))\mathrm{d}x.
    \end{aligned}
    \end{equation*}
    The variation of $\mathcal{L}$ with respect to $u_{\hat{i}}$:
    \begin{equation*}
        \frac{\partial\mathcal{L}}{\partial u_{\hat{i}}}=r_{\hat{i}}+\varepsilon\ln{u_{\hat{i}}}+\varepsilon-v=0.
    \end{equation*}
    therefore, by the first order optimization condition, we have 
    $$u^{t+1}_{\hat{i}}(x)=e^{\frac{-r_{\hat{i}}(x)+v(x)}{\varepsilon}-1},$$
    Furthermore, sum both sides of the equation and use the constraint condition, one can obtain
    $$1=\sum_{\hat{i}=1}^Iu_{\hat{i}}(x)=e^{\frac{v(x)}{\varepsilon}-1}\sum_{\hat{i}=1}^I{e^{\frac{-r_{\hat{i}}(x)}{\varepsilon}}},$$
    Substitute the above expression back into the expression for $u^{t+1}_{\hat{i}}$, finally we obtain
    \begin{equation*}
        \!u^{t+1}_{\hat{i}}=\!\frac{e^{\frac{-r_{\hat{i}}}{\varepsilon}}}{\sum\limits_{i^{'}=1}^{I}e^{\frac{-r_{i^{'}}}{\varepsilon}}}\!=\!\frac{e^{\frac{\bm{o}_{\hat{i}}-p_{\hat{i}}^t-\delta_{\hat{i},i}\text{div}(q^{t+1}\bm{T}^t)}{\varepsilon}}}{\sum\limits_{i^{'}=1}^{I}e^{\frac{\bm{o}_{i^{'}}-p_{i^{'}}^t-\delta_{i^{'},i}\text{div}(q^{t+1}\bm{T}^t)}{\varepsilon}}},\hat{i}=1,\dots,I.
    \end{equation*}%
    }
\section{Hyperparameter Impact}
\label{appendix_C}
\begin{figure}[htbp]
        \centering
		\subfloat[\small {Varying $\lambda$} ]{\includegraphics[width=0.235\linewidth]{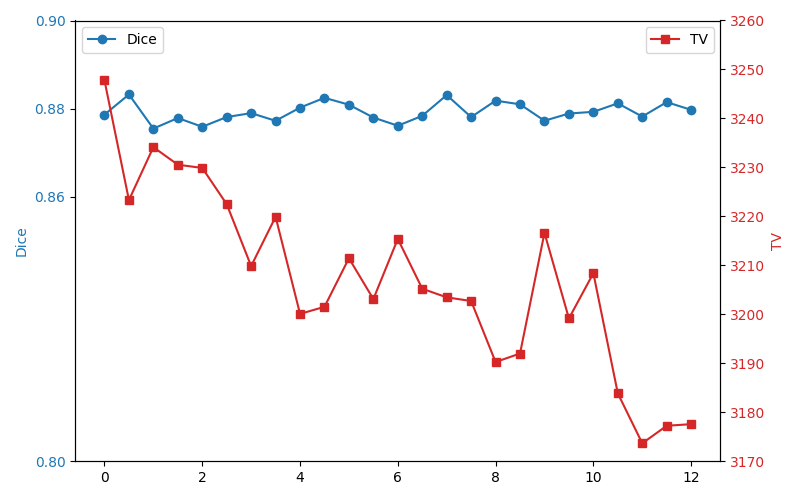}
        \label{lambda}}
        \subfloat[\small {Varying $\varepsilon$} ]{\includegraphics[width=0.235\linewidth]{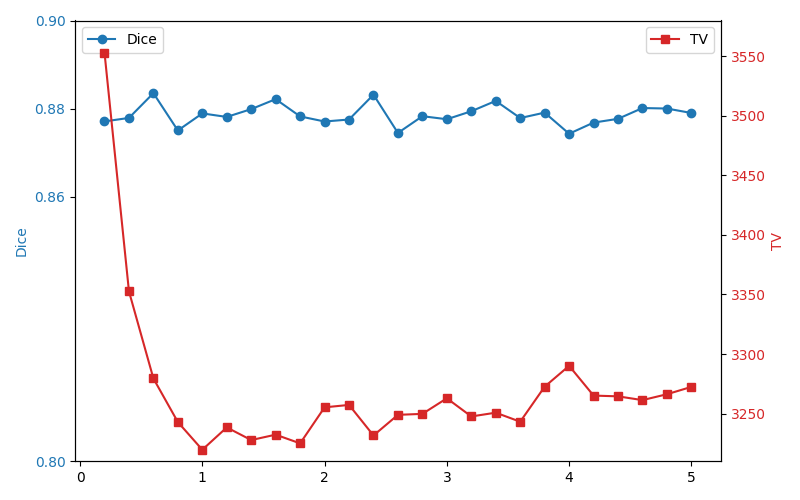}
        \label{varepsilon}
        }
        \subfloat[\small {Varying $\tau_q$}]{\includegraphics[width=0.235\linewidth]{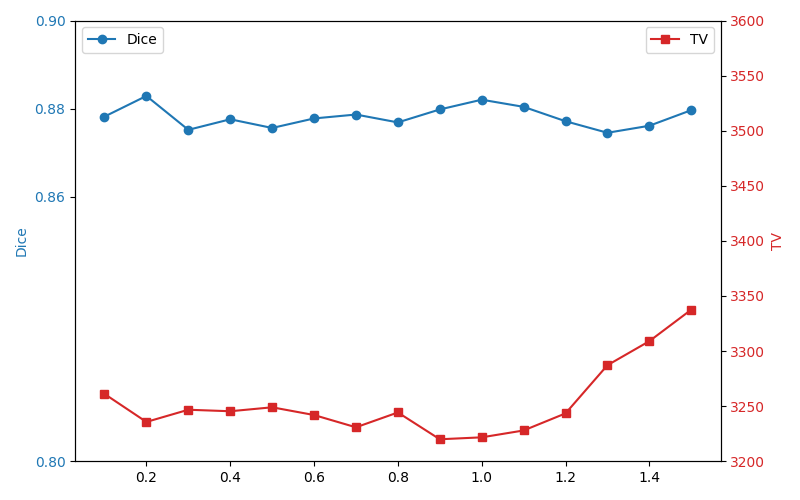}
        \label{tauq}
        }
        \subfloat[\small {Varying $\sigma$}]{\includegraphics[width=0.235\linewidth]{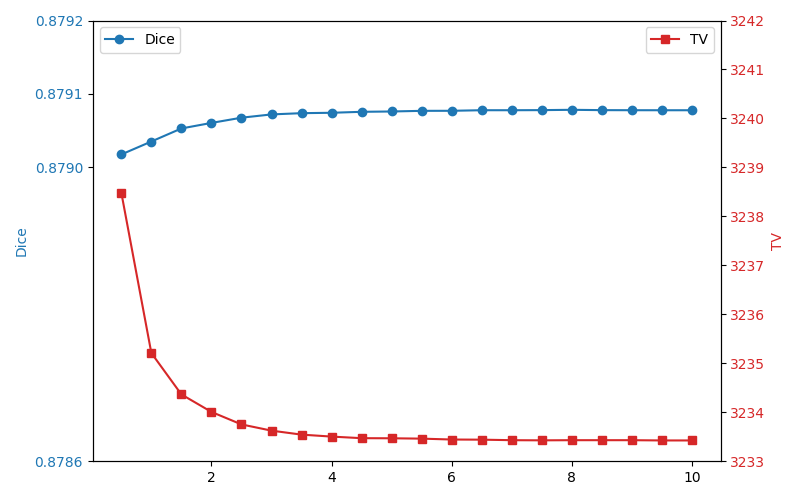}
        \label{sigma}
        }
        \caption{Dice score (blue) and TV value (red) on the REFUGE test set when varying (a) $\lambda$, (b) $\varepsilon$, (c) $\tau_q$ and (d) $\sigma$.}
        \label{fig:hyperpara}
    \end{figure}

{We systematically evaluated the impact of the hyperparameters $\lambda$, $\varepsilon$, $\tau_q$ and $\sigma$ in SAM-post. For each experiment, we varied one parameter while fixing the others to their default values($\lambda=1$, $\varepsilon$=1, $\tau_q$=1, $\sigma=5$). We then measured the Dice score and total variation (TV, which quantifies segmentation smoothness) on the REFUGE test set.  Both the regularization weight $\lambda$ and entropy weight $\varepsilon$ mainly affect smoothness: larger values produce smoother segmentations, reflected by decreasing TV, while the Dice score remains nearly unchanged (\figurename\ref{lambda}, \figurename\ref{varepsilon}). 
 {In addition, the Gaussian kernel parameter $\sigma$, which governs the spatial extent of pixel-level smoothing, also contributes to segmentation smoothness: as $\sigma$ increases, the TV value decreases and the Dice score slightly increases, yet the magnitudes of these changes are relatively small (\figurename\ref{sigma}).} 
The step size $\tau_q$ controls the update of dual variables. An excessively large value may lead to divergence, while within a reasonable range it has little impact on the final results (\figurename\ref{tauq}).}
\section{Experiments based on other models}
\label{appendix_D}
\begin{figure}[htbp]
		\centering
        \captionsetup[subfloat]{labelformat=empty}
		\subfloat[\small Image]{%
			\begin{minipage}[b]{0.23\linewidth}
				\includegraphics[width=\linewidth]{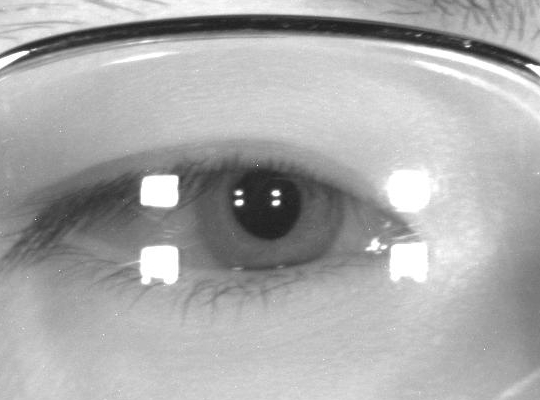}\vspace{1pt}
                \includegraphics[width=\linewidth]{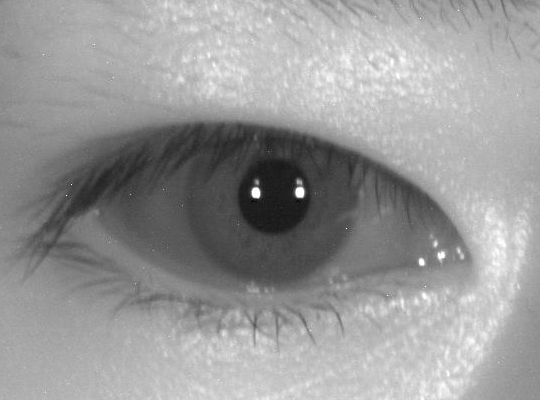}
			\end{minipage}%
            \vspace{-4pt}
	}%
		\subfloat[\small Ground Truth]{%
			\begin{minipage}[b]{0.23\linewidth}
				\includegraphics[width=\linewidth]{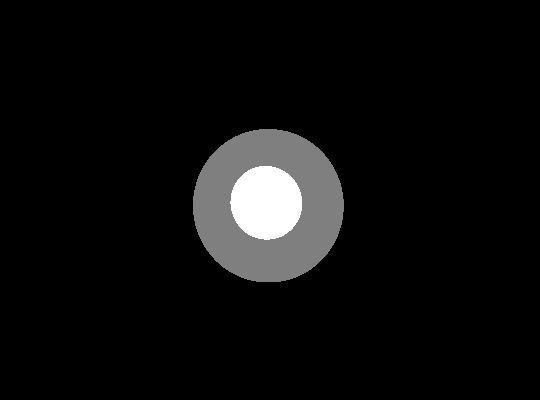}\vspace{1pt}
				\includegraphics[width=\linewidth]{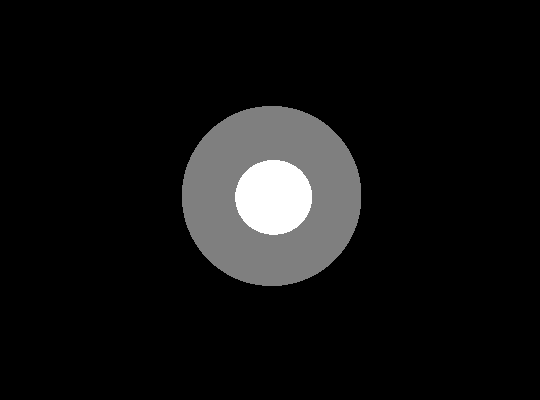}
			\end{minipage}%
            \vspace{-4pt}
	}%
		\subfloat[\small Unet++]{%
			\begin{minipage}[b]{0.23\linewidth}
				\includegraphics[width=\linewidth]{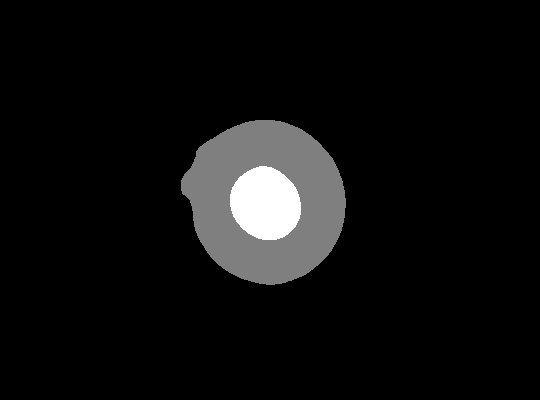}\vspace{1pt}
				\includegraphics[width=\linewidth]{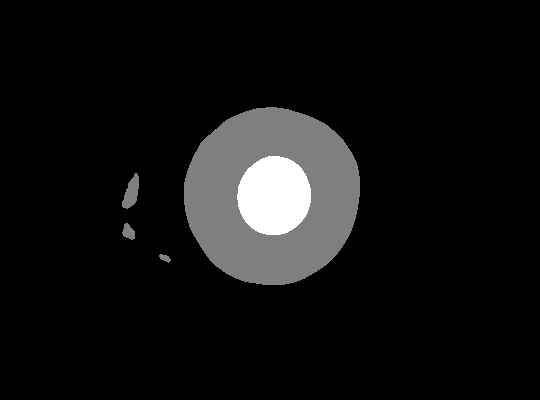}
			\end{minipage}%
            \vspace{-4pt}
		}%
		\subfloat[\small +ESP]{%
			\begin{minipage}[b]{0.23\linewidth}
				\includegraphics[width=\linewidth]{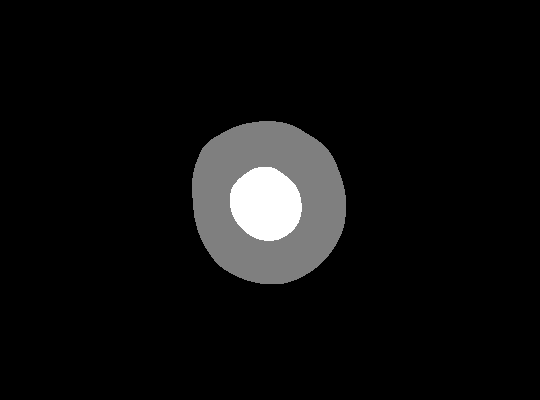}\vspace{1pt}
				\includegraphics[width=\linewidth]{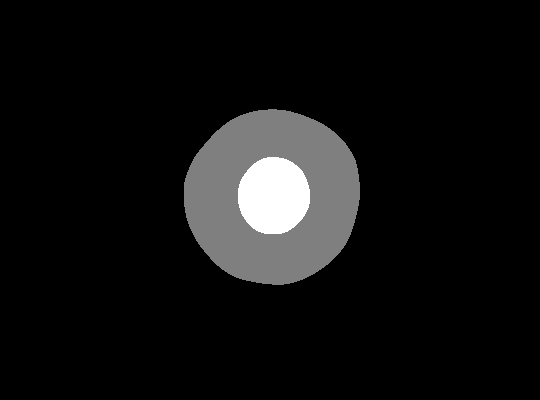}
			\end{minipage}%
            \vspace{-4pt}
		}%
		\caption{Visualization of segmentation results on CASIA.v4 datasets.}
		\label{fig:unet_model}
	\end{figure}

 {
In this section, we conduct experiments using Unet++ as baseline models to validate the effectiveness of the proposed Ellipse Shape Prior (ESP) module on the CASIA.v4 dataset. The network is trained to simultaneously segment both pupil and iris regions, with unified ellipse shape constraints applied to these two elliptical structures. Quantitative results presented in Table \ref{result-other} demonstrate that incorporating the ESP module improves the Dice scores for both categories, with particularly significant enhancements in boundary accuracy metrics BD and BDSD. Qualitative results illustrated in \figurename \ref{fig:unet_model}  further confirm the module's capability to enhance the segmentation of elliptical structures.
\begin{table}[htbp]
	\centering
	\caption{Test Results based on Other Model}
	\label{result-other}
	\begin{threeparttable}
		\begin{tabular}{lc|ccc}
			\hline
			\hline  
          {Class}&{Network} & {Dice$\uparrow$} & {BD$\downarrow$} & {BDSD$\downarrow$}\\
         \hline
         \multirow{2}{*}{pupil}&{Unet++} & {94.19$\pm$0.42} & {7.97$\pm$2.36} & {10.24$\pm$3.14}\\
         {}&{+ESP} & {\textbf{94.57}$\pm$0.06} & {\textbf{2.82}$\pm$0.91} & {\textbf{3.21}$\pm$1.93}\\
         \hline
         \multirow{2}{*}{iris}&{Unet++} & {95.36$\pm$0.13} & {8.75$\pm$1.78} & {11.99$\pm$1.78}\\
         {}&{+ESP} & {\textbf{95.43}$\pm$0.18} & {\textbf{6.53}$\pm$1.10} & {\textbf{8.65}$\pm$1.61}\\
         \hline
         \hline
		\end{tabular}
	\end{threeparttable}
\end{table}	
}


\bibliographystyle{IEEEtran}
\bibliography{reference}
\begin{IEEEbiography}
[{\includegraphics[width=1in,height=1.25in,clip,keepaspectratio]{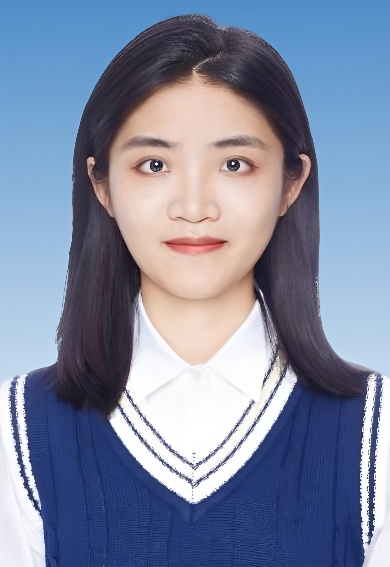}}]{Xinyu Zhao}
received the B.S. degree in Information and Computational Science from Beijing Institute of Technology in 2023. She is currently pursuing
the M.S. degree at Beijing Normal University. Her
research interests include variational image processing,
deep learning and its applications in image segmentation.
\end{IEEEbiography}
\vspace{-8mm}
\begin{IEEEbiography}[{\includegraphics[width=1in,height=1.25in,clip,keepaspectratio]{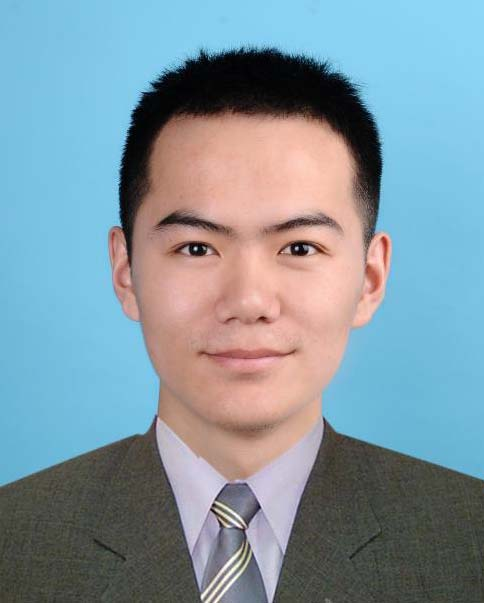}}]{Faqiang Wang}
received the B.S. degree in mathematics and applied mathematics from Heilongjiang University, China in 2014. He received the M.S. degree in computational mathematics and the Ph.D. degree in applied mathematics from Beijing Normal University (BNU), China in 2017 and 2020 respectively. He is currently a lecturer at BNU. His main research interests include variational and PDE-based image processing, optimal transport and deep learning based methods.
\end{IEEEbiography}
\vspace{-8mm}
\begin{IEEEbiography}[{\includegraphics[width=1in,height=1.25in,clip,keepaspectratio]{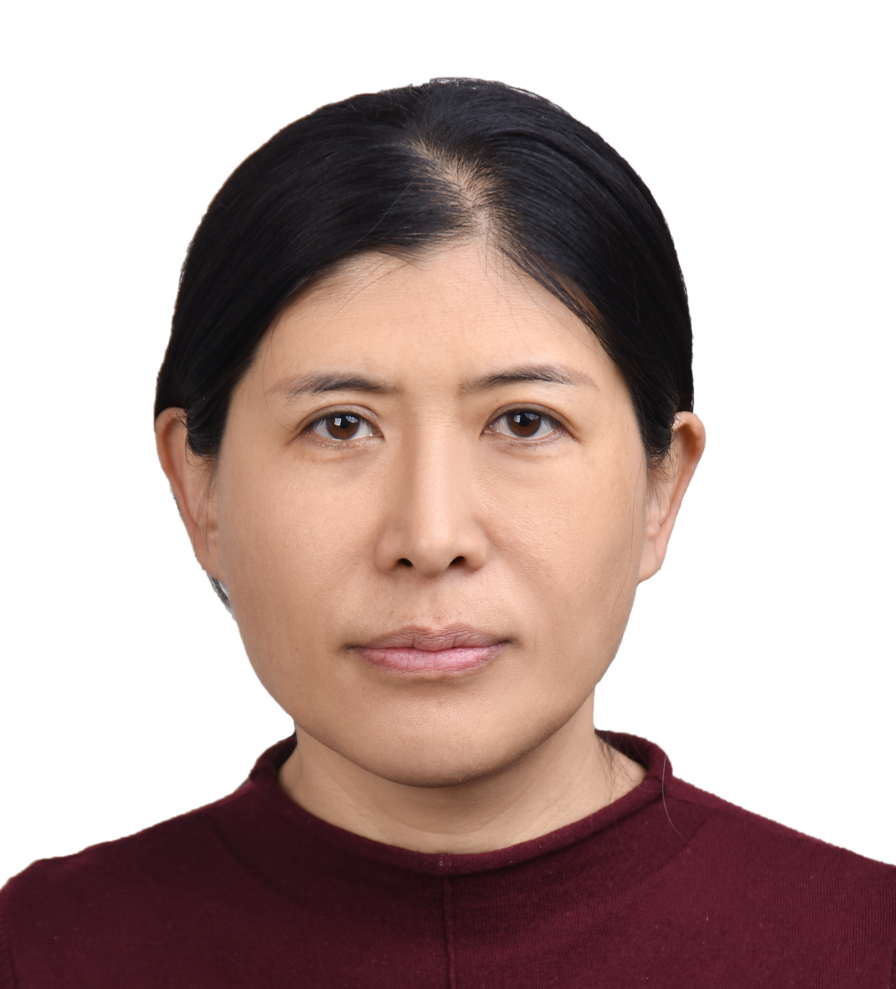}}]{Li Cui}
received the Ph.D. degree in computational mathematics from Jilin University, Changchun, China, in 2004. She is currently an associate professor with the School of Mathematical Sciences, Beijing Normal University. Her main research interests include optimal transportation theory and its application in deep learning.\relax
\end{IEEEbiography}
\vspace{-8mm}
\begin{IEEEbiography}[{\includegraphics[width=1in,height=1.25in,clip,keepaspectratio]{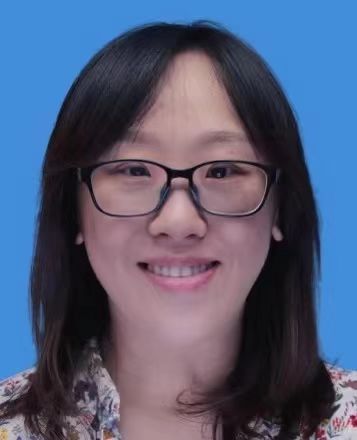}}]{Yuping Duan}
is a full professor at the School of Mathematical Sciences of Beijing Normal University (BNU). Before joining BNU, she was a professor at Tianjin University in 2015 to 2023, and a research scientist at I2R, A*STAR in 2012 to 2015. She received her Ph.D. from Nanyang Technological University in 2012. Her research interests are image processing and computer vision, variational methods, and deep learning methods.
\end{IEEEbiography}
\vspace{-8mm}
\begin{IEEEbiography}
[{\includegraphics[width=1in,height=1.25in,clip,keepaspectratio]{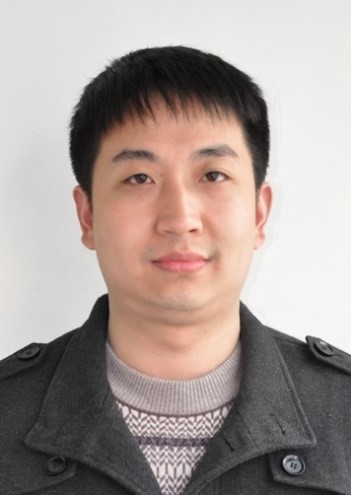}}]{Jun Liu}
received the B.S. degree in mathematics from the Hunan Normal University, China in 2004. He received the M.S. and Ph.D. degrees in computational mathematics from the Beijing Normal University (BNU), China, in 2008 and 2011 respectively. He is currently a professor at BNU. His research interests include variational, optimal transport and deep learning based image processing algorithms and their applications.
\end{IEEEbiography}

\end{document}